%% file: main.tex
\documentclass{article}

\usepackage[accepted]{icml2024}
\usepackage{microtype}
\usepackage{graphicx}
\usepackage{subfig}
\usepackage{booktabs} %
\usepackage{bbm}
\usepackage{nicefrac}

\usepackage[utf8]{inputenc}
\usepackage[T1]{fontenc}

\usepackage{hyperref}

\usepackage{icml2024}

\usepackage{amsmath}
\usepackage{cancel}
\usepackage{amssymb}
\usepackage{mathtools}
\usepackage{amsthm}
\usepackage{threeparttable}
\usepackage[normalem]{ulem} %
\usepackage{siunitx}
\usepackage{ amssymb }

\usepackage[capitalize,noabbrev]{cleveref}

\theoremstyle{plain}
\newtheorem{theorem}{Theorem}[section]

\theoremstyle{definition}
\newtheorem{definition}[theorem]{Definition}

\theoremstyle{remark}

\usepackage[textsize=tiny]{todonotes}
\usepackage{subfiles} %

\icmltitlerunning{Going beyond Compositions, DDPMs Can Produce Zero-Shot Interpolations}
\usepackage{accents}

\DeclareMathOperator{\preceps}{\precsim_\epsilon}
\DeclareMathOperator{\lbz}{\underaccent{\bar}{z}}
\DeclareMathOperator{\ubz}{\bar{z}}

\newcommand{\imgsizeusedsmall}{$64\times64$}

\newcommand{\rebuttal}[1]{\textcolor{black}{#1}}

\begin{document}

\twocolumn[
\icmltitle{Going beyond Compositions, DDPMs Can Produce Zero-Shot Interpolations}

\icmlsetsymbol{equal}{*}

\begin{icmlauthorlist}
\icmlauthor{Justin Deschenaux}{equal,ic}
\icmlauthor{Igor Krawczuk}{equal,ee}
\icmlauthor{Grigorios G. Chrysos}{wisc,ee}
\icmlauthor{Volkan Cevher}{ee}
\end{icmlauthorlist}
\icmlaffiliation{ic}{Department of Computer Science, École Polytechnique Fédérale de Lausanne (EPFL), Lausanne, Switzerland}
\icmlaffiliation{ee}{LIONS, École Polytechnique Fédérale de Lausanne (EPFL), Lausanne, Switzerland}
\icmlaffiliation{wisc}{Department of Electrical and Computer Engineering, University of Wisconsin-Madison, USA}

\icmlcorrespondingauthor{Justin Deschenaux}{justin.deschenaux@gmail.com}
\icmlcorrespondingauthor{Igor Krawczuk}{igor@krawczuk.eu}

\icmlkeywords{Machine Learning, ICML}

\vskip 0.3in
]

\printAffiliationsAndNotice{\icmlEqualContribution} %

\begin{abstract}
 Denoising Diffusion Probabilistic Models (DDPMs) exhibit remarkable capabilities in image generation, with studies suggesting that they can generalize by composing latent factors learned from the training data. In this work, we go further and study DDPMs trained on strictly separate subsets of the data distribution with large gaps on the support of the latent factors. We show that such a model can effectively generate images in the unexplored, intermediate regions of the distribution. For instance, when trained on clearly smiling and non-smiling faces, we demonstrate a sampling procedure which can generate slightly smiling faces without reference images (zero-shot interpolation). We replicate these findings for other attributes \rebuttal{as well as other datasets. \href{https://github.com/jdeschena/ddpm-zero-shot-interpolation}{Our code is available on GitHub}.}
 \end{abstract}

\subfile{sections/introduction}

\subfile{sections/background}

\subfile{sections/related_work}

\subfile{sections/method}

\subfile{sections/experiments}

\subfile{sections/conclusion}
\newpage
\subfile{sections/impact_statement}

\subfile{sections/acknowledgements}

\bibliography{main}
\bibliographystyle{icml2024}

\newpage
\appendix
\renewcommand{\thefigure}{S\arabic{figure}} %
\renewcommand{\thetable}{S\arabic{table}} %
\renewcommand{\thetheorem}{S\arabic{theorem}} %
\onecolumn
\subfile{sections/appendix}

\end{document}

%% file: sections/introduction.tex
\section{Introduction}

\begin{figure}[t] 
    \centering
    \includegraphics[width=1\linewidth]{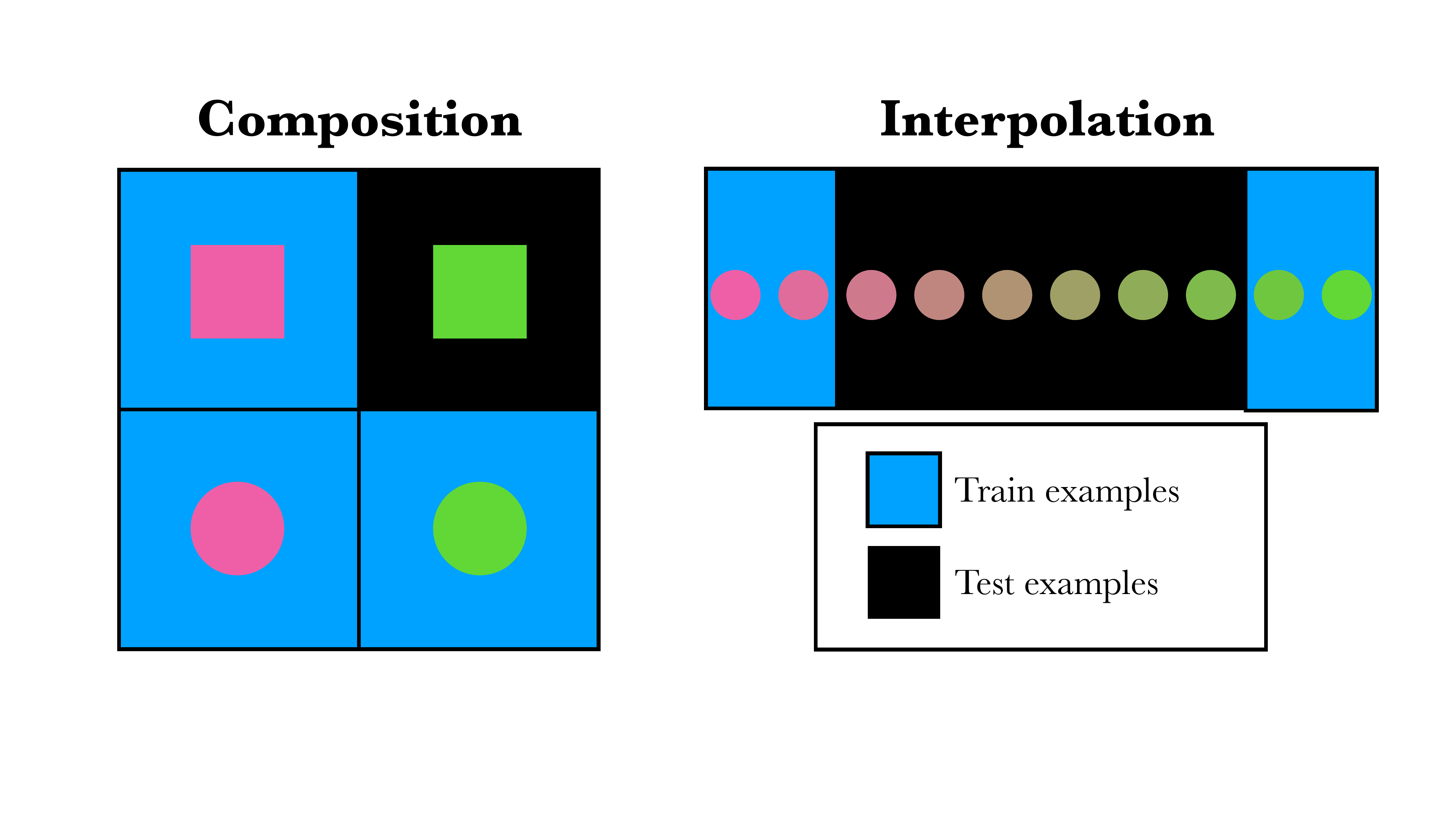}
    \vspace{-15mm}
    \caption{\rebuttal{Assume that the characteristic of real samples is influenced by some latent variables $z_i$ (e.g. color and shape). If the support of the latent space is a cartesian product of the individual latent variables (e.g. pairs (color, shape) typically finite and discrete), we say that the samples depend on \textbf{combinations of the individual latent variables}. Conversely, if the latent variables $z_i$ are defined on a closed interval (e.g. $z_i \in [0, 1]$), such that any value of $z_i$ induces a meaningful sample (e.g. any color in between green and pink), then we say that there exists an \textbf{interpolation between extreme samples}, where the extreme examples are samples whose latent are close to the extrema of their support (e.g. pure pink or pure green colors).}}

    \label{fig:interpolation-vs-composition}
    \vspace{-5mm}
\end{figure}
In recent years Diffusion models (DM) \citep{sohldickstein2015deep, ddpm, song2021scorebased} have emerged as highly successful generative methods for content synthesis \citep{improved-ddpm,
rombach2022highresolution,
du2020compositional}.
As observed by \citet{okawa2023compositional} and others, DMs are capable of leveraging \emph{compositionality}, an important concept in psychology \citep{frege1980philosophical}%
, philosophy \citep{Pelletier1994} %
and linguistics \citep{williams1983semantic}
 which refers to the human capability to combine attributes into novel configurations.
For instance, after observing smiling people with both long hair and frowning people with short hair, humans can imagine smiling short-haired individuals.
Furthermore, humans are able to \emph{interpolate} attributes as well, e.g. they will be able to imagine short-haired individuals with mild smiles or neutral expressions as well.
Crucially, interpolating implicitly assumes there exists a latent factor governing a set of attributes, inducing a natural ranking of samples according to the factor of interest. In the sequel, we will refer to attributes as the \emph{expressions} corresponding to a latent variable which induces a ranking. \textbf{Remarkably, \rebuttal{we observe that} diffusion models are capable of interpolation without being trained on instances with intermediate expressions of the latent factor}. In the example, a generative model capable of interpolation would not be exposed to neutral individuals or mild smiles and frowns, yet could generate samples exhibiting these facial expressions.

Interpolation abilities are crucial to address questions related to fairness and bias mitigation in machine learning through generative models \cite{8869910, 10.1007/978-3-030-90963-5_39, tan2021improving}.
Indeed, a model able to interpolate would allow generating nuanced samples even when the training dataset contains little diversity.
This is especially important because the data collection process might filter out instances that do not meet ad-hoc criteria. Such procedure would result in unexpectedly non-representative training sets for specific underrepresented attributes.
In addition, a model with strong interpolation would prove invaluable in scenarios where specific expressions of the latent factor of interest can only be sparsely observed.
For instance, in medical imaging, where DMs are being explored as priors for MRI reconstruction\citep{GUNGOR2023102872}, certain rare pathologies may not be sufficiently represented. Thus the capacity of the model to seamlessly interpolate is crucial for generating (and accurately reconstructing) diverse data.

While the ability to compose fixed attributes was identified in diffusion models in controlled\footnote{Production text-to-image models deployed in applications use vast amounts of web-scraped training data. Hence, Occam's razor might suggest that novelties emerge as compositions of latent factors from the training data.} settings \citep{okawa2023compositional}, this work presents, for the first time, a clear demonstration that \rebuttal{Denoising Diffusion Probabilistic Models} (DDPM) can reliably interpolate between training classes, going far beyond the support of the training data, as we illustrate in \cref{fig:interpolation-vs-composition}.

In particular, we demonstrate zero-shot interpolation of DDPMs, wherein all models are exclusively trained on data clearly excluding intermediate values of the latent factor. \rebuttal{We now present our main contributions. For reasons of space, we defer certain results to the appendix:}

\begin{itemize}
    \item We demonstrate the interpolation abilities of DDPMs trained on examples with one factor of interest in \cref{sec:interpolation-realizable}. We demonstrate this ability on real-world datasets, filtered to retain examples with clear attributes only, as depicted in \cref{fig:interpolation-vs-composition} (right). Importantly, the training samples are highly separated in our experiments.
    \item We demonstrate that interpolation can happen in low data regimes in \cref{sec:how-little-data-can-we-use}, is robust to hyperparameter choices in \cref{sec:sensitivity-to-guidance}, \rebuttal{model architecture, and occurs with various sampling strategies (\cref{sec:interpolation-with-different-hyperparameters})}. We further explore regularization to improve interpolation capabilities in \cref{sec:interp-toy-dataset} 
    \item \rebuttal{We demonstrate that DMs trained on clear attributes only can perform attribute editing, and generate images beyond the training distribution (\cref{sec:attr-edit-without-mild})}.
    \item \rebuttal{In \cref{sec:interpolation-ddim-latents}, we demonstrate that decoding interpolated latents generates samples with intermediate expression of the latent factors, despite being not trained on such examples.}
    \item \rebuttal{Finally}, we present interpolation on two attributes in \cref{sec:interpolation-two-variables}.
\end{itemize}

%% file: sections/background.tex
\section{Background}
\label{sec:background}
Consider statistical models that approximate a distribution $p(X, y)$, where $X \in \mathbb R^D$ represents high-dimensional signals such as images, and $y \in \mathbb R^L$ encodes conceptual information about the image.
In recent years, diffusion models have garnered significant attention in the research community for their effectiveness in modeling and generating approximate samples from $p(X, y)$.
\begin{figure}[t] 
    \centering
    \includegraphics[width=0.24\linewidth]{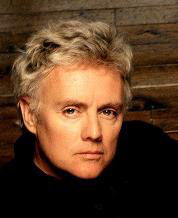}
    \includegraphics[width=0.24\linewidth]{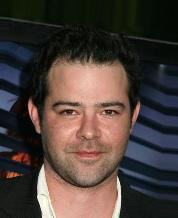}
    \includegraphics[width=0.24\linewidth]{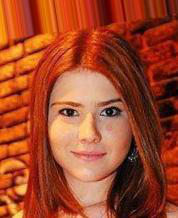}
    \includegraphics[width=0.24\linewidth]{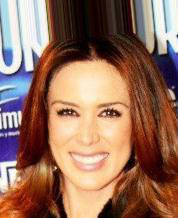}
    \vspace{-2mm}
    \caption{Images from CelebA dataset. \textbf{Left}: clearly non-smiling face. \textbf{Two center}: mild smiles. \textbf{Right}: clearly smiling.}
    \label{fig:smile-non-smile-mild-celeba}
    \vspace{-5mm}
\end{figure}
\subsection{Diffusion and Energy-Based Models}
\label{sec:diffusion-models}
Introduced in \citet{sohldickstein2015deep}, diffusion models (DMs) serve as the foundation of many state-of-the-art text-to-image models \citep{ramesh2022hierarchical, rombach2022highresolution, saharia2022photorealistic}. \citet{ddpm} introduced Denoising Diffusion Probabilistic Models (DDPMs), and demonstrated that diffusion models could match the FID of GANs for certain datasets. 

DDPMs materialize the diffusion process as a hierarchy of $T$ increasingly noisy samples, mapping the original distribution to white noise, \rebuttal{where} sampling is achieved through variants of the stochastic gradient Langevin dynamics algorithm \citep{welling-langevin-sampling}. The noise addition process is governed by a schedule $(\alpha_t)_{t = 1}^T$, such that $p(x_t | x_{t - 1}) \sim \mathcal N \left(x_t; \sqrt{\alpha_t} x_{t - 1}, (1 - \alpha_t) \mathbf{I} \right)$. Because of the Gaussian transition, one can compute $x_t | x_0$ in a single step, as $x_t = \sqrt{\bar \alpha_t}x_0 + \sqrt{1 - \bar \alpha_t} \epsilon$, where $\bar \alpha_t = \prod_{i=1}^t \alpha_i$, and $\epsilon \sim \mathcal N(0, \mathbf{I})$. \citet{ddpm} sample from the data distribution $p_\text{data}$ by approximating the denoising distribution $p(x_{t - 1} | x_t )$ with a Gaussian $p_\theta(x_{t - 1} | x_t ) = \mathcal N \left(x_{t-1} | \mu_\theta(x_t, t); \sigma_t \mathbf{I} \right)$, where $\sigma_t$ is computed in terms of $(\alpha_t)_{t = 1}^T$ and $\mu_\theta(x_t, t) = \frac{1}{\sqrt{\alpha_t}}\left(x_t - \frac{1 - \alpha_t}{\sqrt{1 - \bar \alpha_t}}\epsilon_\theta(x_t, t) \right)$, where $\epsilon_\theta(x_t, t)$ estimates the added noise. \citet{ddpm} train $\epsilon_\theta$ to minimize $L_\text{simple} := \mathbb E_{t, x_0, \epsilon} \left[ \| \epsilon - \epsilon_\theta (x_t, t) \|^2 \right]$, where $t$ is picked uniformly at random in $1, ..., T$, $\epsilon \sim \mathcal N (0, \mathbf I)$, and $x_0$ is a data sample. See \cref{appendix:diffusion-models} for more details.

Energy-based models (EBMs) \citep{ngiam2011learning, xie2016theory, NEURIPS2019_378a063b, du2020implicit, du2021improved} parameterize a distribution $p_\theta(x)$ through the non-normalized Energy function $E(\cdot)$, as $p_\theta(x) \propto e^{- E(x)}$. Notably, the energy of a product distribution is the sum of energies of each component. EBMs and diffusion models share functionally equivalent learning objective and sampling method \citep{liu2023compositional}. Interestingly, \citet{liu2023compositional} osbserved that summing the score from multiple pre-trained diffusion models improves composition ability on CLEVR \citep{johnson2016clevr}.
\subsection{Conditional Generation with Classifier Guidance}
\label{sec:classifier-guidance}
\rebuttal{In the main body of this work, we study conditional generation using classifier guidance \citep{song2021scorebased, dhariwal2021diffusion} as it lets us reuse unconditional models during our explorations. Since classifier-free guidance (CFG) \citep{ho2022classifierfree} typically achieves superior FID \citep{ttur-gan-fid} and IS \citep{salimans2016improved}, we study CFG and present results in \cref{sec:interpolation-with-different-hyperparameters}. We observe interpolation capabilities with both sampling schemes.}
Practically, DDPMs implement classifier guidance by modifying the mean of the denoising distribution $p_\theta (x_{t - 1} | x_t)$ from \cref{sec:diffusion-models}. Formally, the new mean is computed as $\tilde \mu_\theta = \mu_\theta(x_t, t) + \sigma_t^2 g$, where $g = \lambda \nabla_{x} \log p_\theta(y | x)|_{x = \mu_\theta (x_t, t)}$, $\lambda \in \mathbb R$ (guidance strength) is such that $|\lambda| \geq 1$ and $p_\theta(y | x)$ is learned with a classifier.
\subsection{Compositional Generalization}
\label{sec:compositional-generalization}
We build upon the \rebuttal{concept} of compositional generalization of \citet{wiedemer2023compositional} to define interpolation in \cref{sec:problem-statement}. Compositional generalization denotes the ability to compose attributes in novel ways.
Importantly \citet{wiedemer2023compositional}  introduce the concept of \emph{compositional support}:
\begin{definition}[\citet{wiedemer2023compositional}] 
    Let $P$ and $Q$ be distributions defined on a latent domain $\mathcal Z_1 \times ... \times \mathcal Z_K$. We say that $P$ and $Q$ have \textbf{compositional support} when $P$ and $Q$ have matching marginals' support. Formally, denoting as $P_k$ the marginalization of $P$ over $\mathcal Z_i \neq \mathcal Z_k, ~\forall i \neq k$, compositional support between $P$ and $Q$ is achieved when
    \begin{equation}
        \text{supp} ~ P_k = \text{supp} ~ Q_K, \quad \forall k = 1,...,K,
    \end{equation}
    where $\text{supp}$ denotes the support of a distribution.
    \label{def:compositional-generalization}
\end{definition}
\citet{wiedemer2023compositional} argue that compositional generalization (sampling from $Q$ when trained on $P$) is only possible if the train and test distribution have compositional support.
\citet{okawa2023compositional} suggest that diffusion models have compositional generalization abilities.
\emph{Interpolation, formally defined in \cref{sec:problem-statement}, is distinct from composition}, since the train and test distributions do not have compositional support.
\cref{fig:interpolation-vs-composition} shows visually why interpolation is conceptually independent and separate from composition.

%% file: sections/related_work.tex
\section{Related Work}

\paragraph{Compositionality}
The concept of compositionality broadly implies that a whole can be explained by the sum of its parts, in addition to how components relate to each other.
Compositionality is often attributed to Frege, as developed by \citet{Pelletier1994}.
Prior work in generative modeling have explored compositional aspects of the learned distribution for GANs \citep{chen2016infogan, Shoshan_2021_ICCV}, VAEs \citep{higgins2017betavae, georgopoulos2020multilinear}, diffusion models \citep{liu2023compositional, du2023reduce}, and EBMs \citep{du2020compositional, liu2021learning, du2021unsupervised}.
\citet{wiedemer2023compositional} formally defined compositional generalization and proposed sufficient conditions, but specifically excluded the study of interpolation and extrapolation. Additionally, compositional generalization has been linked to causality \citep{NEURIPS2020_0987b8b3, besserve2021theory}.
Finally, \citet{okawa2023compositional} study diffusion models when the train and test distributions have compositional support, but do not explore interpolation, the focus of our work (\cref{sec:problem-statement}).
\paragraph{Text-to-Image Generation}
In recent years, diffusion models \citep{sohldickstein2015deep, song2020generative, ddpm} showcase impressive abilities to sample high quality compositions based on text prompts \citep{ramesh2022hierarchical, rombach2022highresolution, saharia2022photorealistic}.
Other model classes appear able of composition in this setting as well, eg.. Parti \citep{yu2022scaling}, an autoregressive model, beats Dall·E 2 \citep{ramesh2022hierarchical} and Imagen \citep{saharia2022photorealistic} on FID on MS-COCO \citep{lin2015microsoft}.
On the same dataset, GigaGAN \citep{kang2023gigagan} is also competitive with Stable Diffusion \citep{rombach2022highresolution}.
To be able to cleanly separate interpolation from compositions, we focus in pure vision models and leave the exploration of interpolation in text conditioned and multi-modal models to future work.
\paragraph{Advantages of Diffusion Models}
Diffusion models enjoy stable training, relative ease of scaling \citep{dhariwal2021diffusion}, and offer likelihood estimates of samples \citep{song2021scorebased}. \citet{mallat-geometry} show that models trained on disjoint \rebuttal{sets of samples} can learn the same high-dimensional density, suggesting strong inductive biases.
Closest to our work, \citet{zhu2023unseen} found that DDPMs can perform few-shot non-compositional extrapolation. Our work differs in several key aspects. First, we focus on zero-shot interpolation, whereas \citet{zhu2023unseen} focus on few-shot extrapolation. Additionally, we do not introduce new learnable parameters, while their approach requires fitting high-dimensional Gaussians. 
\paragraph{Diffusion Models and Out-of-Distribution Detection}
In \emph{discriminative} contexts, \citet{graham2023denoising, graham2023unsupervised, liu2023unsupervised} employ diffusion models to identify out-of-distribution (OOD) examples. \citet{du2023dream} employ diffusion models to generate a dataset of outliers.
This dataset is used to train a classifier, as training on borderline outliers enhance OOD detection performance. 
\paragraph{Connections between Diffusion Models and Denoising Score-Matching}
Concurrently to \citet{ddpm, improved-ddpm}, \citet{song2020generative, song2020improved, song2021scorebased} developed analogous formulations related to score matching \citep{10.5555/1046920.1088696, 6795935}. As noted in \citet{du2023reduce}, the DDPMs training objective is proportional to the denoising score-matching objective of \citet{song2020generative}.
\paragraph{Interpolation in Diffusion Models}
\rebuttal{Previous research has explored using diffusion models for interpolation and attribute editing. While impressive, those works do not constrain the training distribution to measure interpolation capabilities. Specifically, \citet{yue2024exploring} present results on attribute editing, leveraging a disentangled representation learned from the entire CelebA training data. \citet{kim2022diffusionclip} conduct attribute editing experiments using pre-trained CLIP models, while \citet{wang2023interpolating} demonstrate latent space interpolation between pairs of images using pre-trained latent diffusion \cite{rombach2022highresolution} and CLIP models.}

%% file: sections/method.tex
\section{Methodology}
\begin{figure*}[t] 
    \centering
    \includegraphics[width=0.6\linewidth]{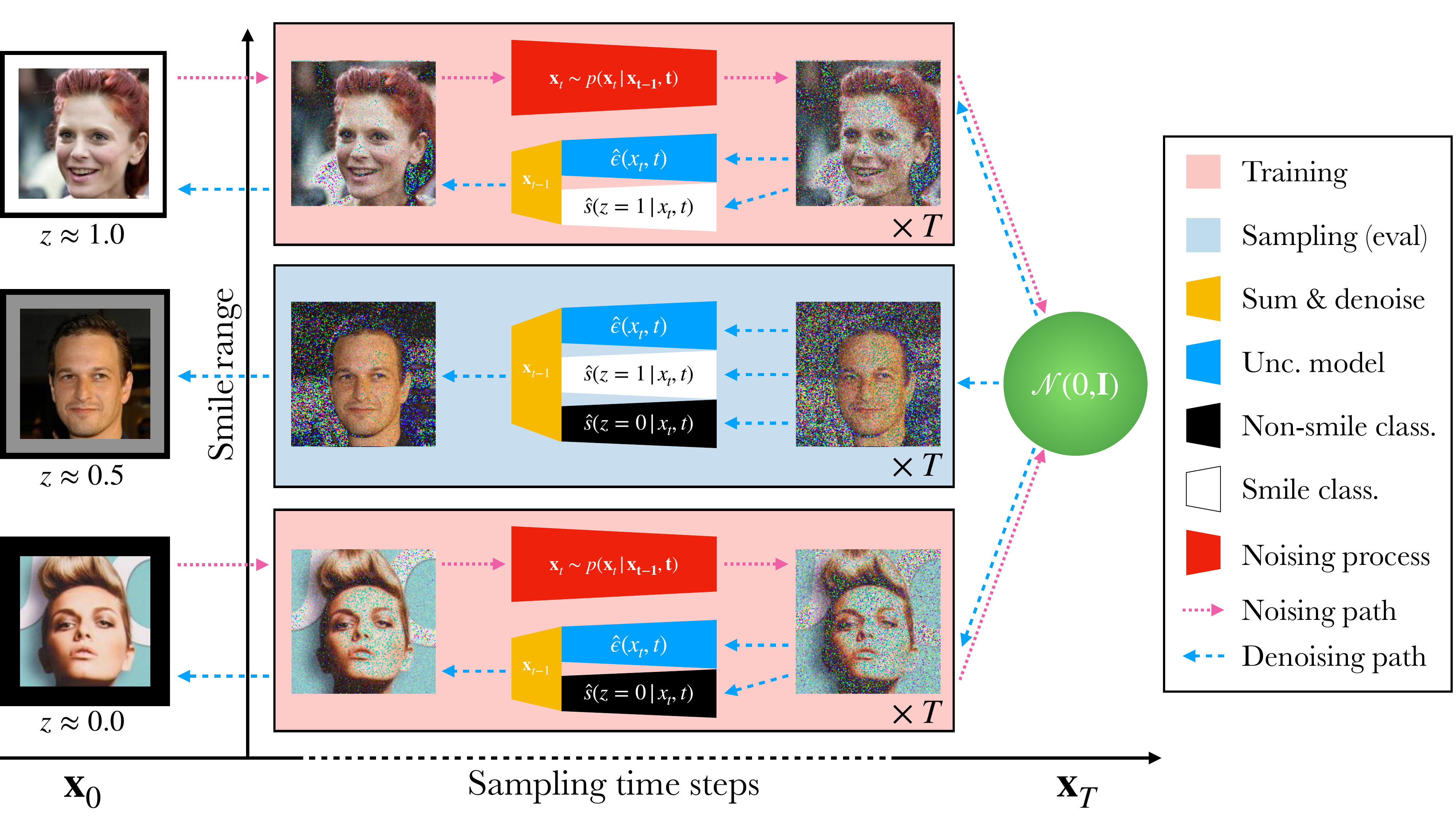}
    \vspace{-3mm}
    \caption{Diagram of the training and sampling process for mild smiles. The classifier and DDPM are trained on \emph{extreme examples only}, i.e. the DDPM is trained on clearly smiling and clearly non-smiling faces. Nonetheless, we demonstrate that DDPMs can generate faces with mild attributes (middle) with a modified sampling scheme, despite \emph{never} encountering those at training. The key for sampling mild attribute is to use the score of the classifier for \emph{both} classes instead of one as in regular \rebuttal{ classifier-guided sampling}. Importantly, we do \emph{not} modify the DDPM training procedure.}
    \label{fig:mild-attribte-sampling}
    \vspace{-3mm}
\end{figure*}
We demonstrate interpolation capabilities of diffusion models beyond the training distribution. In \cref{sec:problem-statement}, we outline the data generation model, present the problem statement and define interpolation formally. We use a guidance mechanism inspired by EBMs, presented in \cref{sec:multi-guidance}.
In \cref{sec:extract-extreme-dataset}, we detail how we extract a subset of examples with clear attributes, required to study interpolation.
\subsection{Problem Statement}
\label{sec:problem-statement}
\paragraph{Data Generation Model}
We denote the set of examples as $\mathcal D := \left\{ (x_i, z_i) : x_i \in \mathbb R^D; z_i \in [0, 1]^L; 1 \leq i \leq N \right\}$, where $N$ denotes the dataset size, $x_i$ represent high-dimensional data such as images, and $z_i$ is a vector of latent variables. Let $z_i(\ell)$ denote the $\ell$-th entry of the vector $z_i$, for $1 \leq \ell \leq L$. Each $z_i(\ell)$ denotes the strength of a certain attribute in sample $x_i$. For example, if $x_i$ is a picture of a face, $z_i(1)$ could model the strength of the smile, where $z_i(1) \approx 0.1$ is supposed to be non-smiling while $z_k(1) \approx 0.9$ means that $x_i$ represents a clearly smiling face. \rebuttal{As explained in the paragraph "Practical considerations", when the dataset does not contain continuous labels, we compute attributes using a classifier. Additionally, if the labels are continuous and take values in $[a, b]$, one can obtain values in $[0, 1]$ via a simple rescaling. We did not observe any drawback to assuming that $z \in [0, 1]$.}
\paragraph{Semi-Order on Attributes}
We assume there exists a ranking between samples based on the latent factor $z_i(\ell)$. 
However, it is often challenging to discern between examples with latent variables having similar values. For instance, when $z_i(\ell)$ represents the smile attribute, distinguishing between $z_i(\ell) = 0.45$ and $z_j(\ell) = 0.5$ visually is challenging, unlike cases with more noticeable differences, such as $z_i(\ell) = 0.05$ and $z_j(\ell) = 0.45$. This notion of minimal distance for comparison is captured by \textbf{semi-orders} \citep{luce1956semiorder,Pirlot1997}. The semi-order $\precsim_\epsilon$ compares samples based on the attribute $z_i(\ell)$, assuming there is a distance $\epsilon > 0$ between latent variables $z_i(\ell)$. Formally, for $(x_0, z_0), (x_1, z_1) \in \mathcal D$, we say that $z_0(\ell) \precsim_\epsilon z_1(\ell)$ if $z_0(\ell) < z_1(\ell)$ \emph{and} $|z_0(\ell) - z_1(\ell)| > \epsilon$. Conversely, when $|z_0(\ell) - z_1(\ell)| \leq \epsilon$, no meaningful ordering can be established between samples based on attribute $\ell$.
\paragraph{Mild and Extreme Attributes}
Let $z(\ell) \in [0, 1]$ be an attribute of interest and $\precsim_\epsilon$ a semi-order on $[0, 1]$. We define \textbf{extreme examples}, as samples with latent variables $z(j) \in [0, \delta] \cup [1 - \delta, 1]$. For a small $\delta$, extreme examples are samples that undoubtedly do or do not have a certain attribute. Conversely, we say that an example has an attribute \textbf{mildly present}, when its associated latent is such that $z(\ell) \in [0.5 - \delta, 0.5 + \delta]$. Importantly, we assume that it is visually possible to tell extreme and mild attributes apart from each other. Formally, given the semi-order precision $\epsilon$, we need that $0.5 - 2 \delta < \epsilon$ so that mild and extreme examples supports are disjoint.
\paragraph{Interpolation without Mild Attributes}
We say that a generative model is able to interpolate, if it is trained on extreme examples only, with a small $\delta$ such as $0.1$ or $0.2$, yet the model can generate examples with mild attributes. Note that samples can be extreme according to one or multiple variables. See \cref{fig:interpolation-vs-composition} (right sub-figure) for a visualization. Demonstrating interpolation requires a clear gap between the training support ($[0, \delta] \cup [1 - \delta, 1]$) and region of mild samples ($[0.5 - \delta, 0.5 + \delta]$).
This requirement stems from the need to distinguish from the findings of \citet{zhao2018bias}, who illustrate that many generative models naturally produce examples slightly different from the training data. In their experiments featuring a toy dataset with colored dots, \citet{zhao2018bias} showcase the ability of the model to generate images with a few more or fewer dots than those present in the training set.
\paragraph{Practical Considerations}
Popular datasets, such as CelebA \citep{liu2015celeba} use binary labels instead of labels with a continuous range. This is not surprising considering the difficulty in ranking all examples with respect to each other. As such, instead of observing latents $z_i(j) \in [0, 1]$, we can only access binary variables $\tilde z_i(j) \in \{0, 1\}$, while the underlying latents are continuous. Despite binary labels, CelebA contains some diversity in attributes, i.e. it is not composed of extreme examples only. Hence, to demonstrate interpolation, we devise a filtering procedure to detect extreme examples in \cref{sec:extract-extreme-dataset}.
\subsection{Multi-Guidance: Sampling using Multiple Scores}
\label{sec:multi-guidance}
In this work we train diffusion models according to the recipe of \citet{improved-ddpm}. That is, we train an unconditional diffusion model, and a guidance classifier on noisy samples. At sampling however, we use a different method, inspired by the connections between EBMs and diffusion models \citep{liu2023compositional}. We define the product distribution $p_\Pi(x)$, as
\begin{equation}
    p_\Pi(x) :=~ \frac{1}{Z} p(x) \prod_{k=1}^K \prod_{m=1}^M p_k(y_m | x)^{\lambda_{km}},
\end{equation}
where $\lambda_{km}$ are classifier-guidance parameters, $Z$ is a normalization constant, $p(x)$ is the unconditional data distribution from an unconditional diffusion model, $p_k(\cdot | x)$ are distributions from guidance classifiers and $y_m$ are the (discrete) values that attribute $k$ can take. As described by \citet{liu2021learning}, sampling from a product distribution can be approximated using the score $\nabla_x \log p_\Pi(x)$, a sum of individual scores from classifiers and unconditional model. Since we train DDPMs, we use $\sum_{k=1}^K \sum_{m=1}^M \lambda_{km} \nabla_x  \log p_k(y_m | x)$ as the guidance score, which generalizes classifier guidance (\cref{sec:classifier-guidance}) to multiple classifiers. We refer to sampling with multiple scores as \textbf{multi-guidance}.
\subsection{Preparing an Extremal Dataset}
\label{sec:extract-extreme-dataset}
Real-world datasets like CelebA, despite their discrete labels, contain some diversity in attributes. For instance, CelebA includes pictures of clearly smiling, non-smiling, and subtly smiling instances (\cref{fig:smile-non-smile-mild-celeba}). To demonstrate interpolation, we first extract extreme examples, in line with standard practice of dataset filtering of \citet{lakshminarayanan2017simple, fang2023data}. More details are given in \cref{appendix:extract-extreme-dataset}. Our procedure relies on two steps:
\textbf{Firstly}, we detect the presence of an attribute using either a pre-trained model or train a model ourselves. Specifically, for the "Young" attribute, we utilize the FaRL model \citep{zheng2022general}, while training EfficientNet classifiers for other attributes. Using the attribute-dependent model, we extract a set $S$ from CelebA training data, where the model is confident and agrees with original labels. \textbf{Secondly}, we train an ensemble of $5$ models on distinct subsets of $S$. We use these models to filter $S$, keeping only examples where all models of the ensemble confidently agree with the labels, resulting in the set $S'$. Furthermore, as an additional precaution, we retain only the $k$ most extreme samples on each side of $[0, 1]$ to compose $S^\star$, the DDPM training data. \rebuttal{In practice, $k$ is in the range $[5000, 60000]$, depending on the experiments. For instance, in \cref{sec:how-little-data-can-we-use}, we study the impact of $k$ on the interpolation capabilities.}
\paragraph{Leakage and Memorization}
We believe our method avoids leakage and present rough estimate of the error probability of the ensemble in \cref{sec:napkin-estimate-filtering-error}. Furthermore, diffusion models can memorize training examples \citep{somepalli2022diffusion, somepalli2023understanding} and reproduce them at inference. Therefore, we report the synthetic examples with closest neighbors in the training data using a CLIP model \citep{radford2021learning} in \cref{sec:memorization-clip}.
Based on these checks, we believe the model does not memorize examples that potentially slipped through the filtering method. Note that we demonstrate interpolation on synthetic datasets in \cref{sec:interp-toy-dataset}, where we can explicitly ensure there are no examples with mild attributes.
In addition to the explicit verification with the CLIP model, interpolation on synthetic data strongly supports our hypothesis that our DDPMs do not interpolate because of leakage or memorization.

%% file: sections/experiments.tex
\begin{figure}[t] 
    \centering
    \includegraphics[width=0.24\linewidth]{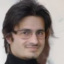}
    \includegraphics[width=0.24\linewidth]{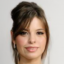}
    \includegraphics[width=0.24\linewidth]{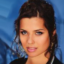}
    \includegraphics[width=0.24\linewidth]{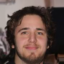}
    \includegraphics[width=0.24\linewidth]{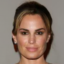}
    \includegraphics[width=0.24\linewidth]{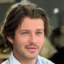}
    \includegraphics[width=0.24\linewidth]{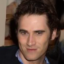}
    \includegraphics[width=0.24\linewidth]{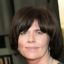}
    \vspace{-2mm}
    \caption{\emph{Synthetic} samples generated with multi-guidance using a DDPM trained on extreme images only. According to the evaluation classifier, the "Smiling" likelihood of the pictures lie in $[0.49, 0.51]$.}
    \label{fig:mildly-smiling-examples}
    \vspace{-3mm}
\end{figure}
\section{Interpolation with One Attribute of Interest}
\label{sec:interpolation-realizable}

In this section, we study the interpolation capabilities of diffusion models, as defined in \cref{sec:problem-statement}. Crucially, we train on $S^\star$, a dataset of \textbf{extreme} examples only. We focus on \imgsizeusedsmall{} images and a single latent factor of interest.
In \cref{sec:smile-interpolation-subsec}, we showcase interpolation results using CelebA's smile attribute \citep{liu2015celeba}.
We reproduce the results for the "Young" and "Blond\_Hair"/"Black\_Hair" attributes in \cref{sec:young-old-interpolation}. Finally, in \cref{sec:interp-toy-dataset}, we demonstrate interpolation for a synthetic dataset of images. More details about the hyperparameters and compute costs in \cref{sec:compute-hparams}. \rebuttal{Although the results in this section use the U-net architecture from \citet{improved-ddpm} with classifier guidance, we still observed interpolation with other architectures and with classifier-free guidance. Further details are provided in \cref{sec:interpolation-with-different-hyperparameters}. Additionally, we present experiments on additional datasets in \cref{sec:object-size-experiment} and \cref{sec:multi-guidance-imagenet}}.

To estimate a proxy for the latent factor values in real-world samples, we train EfficientNet classifiers \citep{tan2021effnet-v2}, following \citet{okawa2023compositional}. \rebuttal{We refer to those classifiers as \textbf{evaluation classifiers}.}
We improve calibration of the classifiers using temperature scaling \citep{guo2017calibration} and use the calibrated probability estimates as our proxy, reported in empirical distribution histograms, \rebuttal{such as \cref{fig:mildly-smiling-histograms}}.
\subsection{Smile Interpolation on CelebA}
\label{sec:smile-interpolation-subsec}
We first focus on the "Smiling" attribute on CelebA as it is relatively balanced, with $48\%$ of \rebuttal{samples} smiling. \rebuttal{Additionally, smiling pictures} can intuitively be ranked according to the smiling strength on a scale.
We train a regular DDPM, and guidance classifier following \citet{improved-ddpm} on a subset of CelebA, filtered as detailed in \cref{sec:extract-extreme-dataset}.

\begin{figure}[t] 
    \centering
    \includegraphics[width=0.45\linewidth]{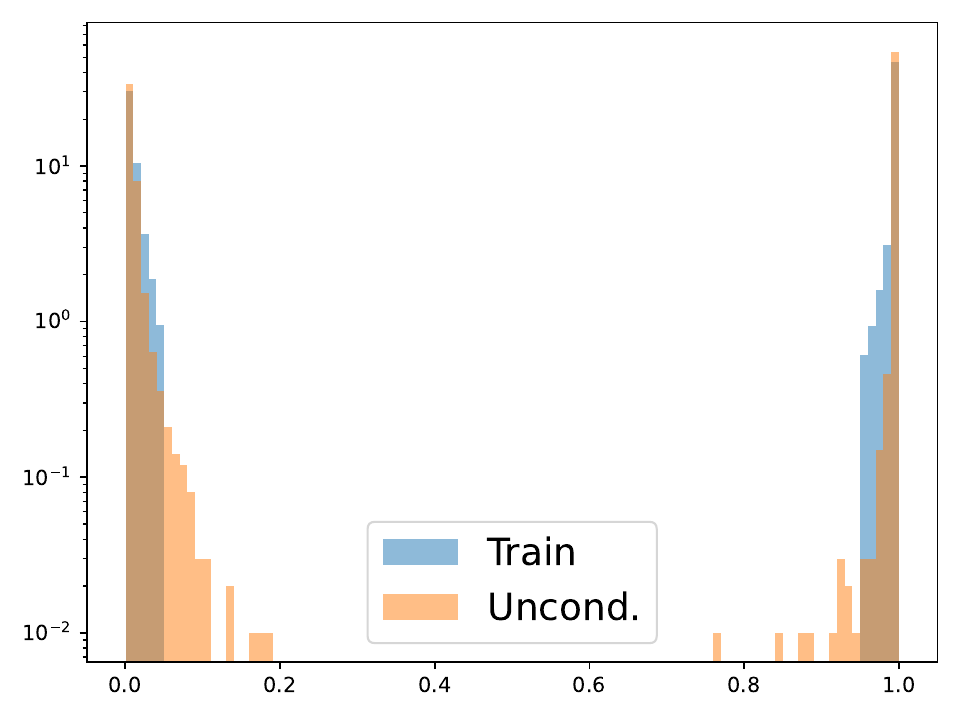}
    \includegraphics[width=0.45\linewidth]{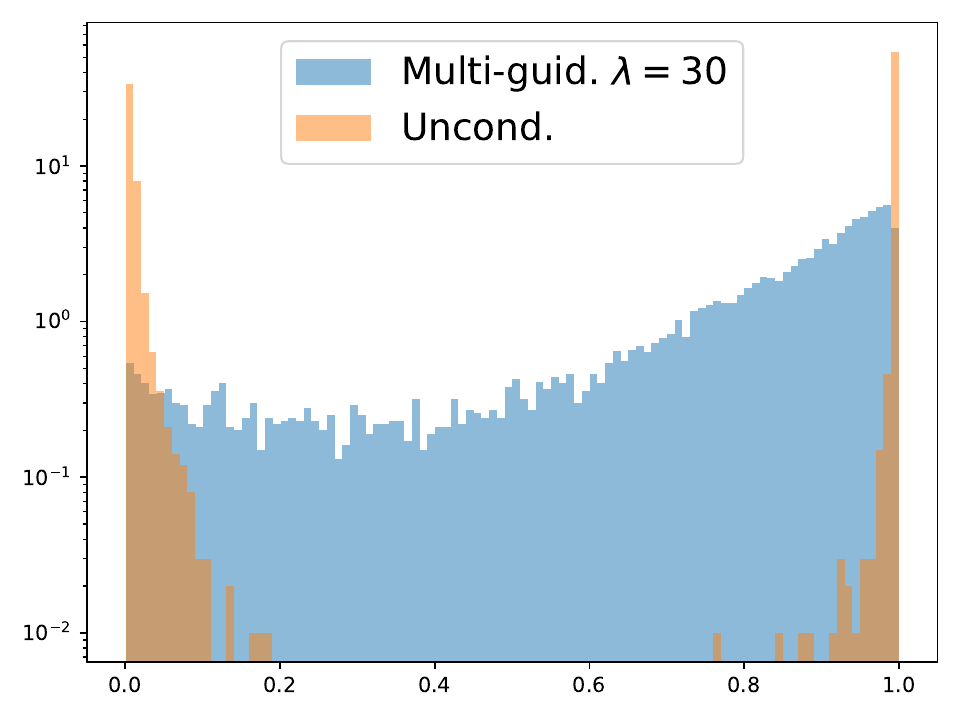}
    \caption{Empirical distribution of pictures according to the evaluation classifier. \textbf{Left}: extreme training examples versus samples from the unconditional diffusion. \textbf{Right}: Samples from the unconditional model versus images sampled with multi-guidance.}
    \label{fig:mildly-smiling-histograms}
\end{figure}

When sampling $10$k examples with multi-guidance ($\lambda_\text{smile} = \lambda_\text{non-smile} = 30$), we observe a distribution significantly closer to uniform than samples from the unconditional model in \cref{fig:mildly-smiling-histograms}. Refer to \cref{fig:mildly-smiling-examples} for mildly smiling examples (\cref{subsec:additional-samples} for more). While the distribution of training examples and unconditional samples is similar, we see that multi-guidance successfully synthesizes examples with mild attributes. For evaluation, we compare the empirical distribution of samples against the uniform distribution. The \textbf{training distribution} has a mean-squared error (MSE) of $0.56$ and Kullback-Leibler divergence (KLD, see \cref{sec:kld-explanation} for details) of $3.17$ (lower is better). The histogram of samples from the \textbf{unconditional model} trained on extreme examples has an MSE of $0.64$ and KLD of $3.5$. Nonetheless, the histogram of samples generated with \textbf{multi-guidance} has an MSE of $0.13$ and KLD of $0.61$. Recall that we check for memorization in \cref{sec:memorization-clip}.%
\subsection{How Little Data Can We Use?}
\label{sec:how-little-data-can-we-use}
We investigate the degradation in performance as we reduce the dataset size. The initial extreme dataset, detailed in \cref{sec:smile-interpolation-subsec}, contains $60$k examples. We create subsets containing $30$k, $10$k, and $5$k examples by ranking all extreme samples with the evaluation classifier and selecting the most extreme instances on each side of the interval. For instance, with a subset of $10$k examples, we choose the $5$k examples with the lowest and highest "Smiling" likelihood. We measure the performance degradation by comparing the empirical distribution against a uniform one. Specifically, we report the MSE and KLD in \cref{tab:smile-data-size-ablation}. See \cref{appendix:smile-data-size-histograms} for histograms. Recall that the training distribution with $60$k samples achieves an MSE of $0.558$ and KLD of $3.161$. While the performance decreases as the data size shrinks, \rebuttal{we observe that DDPMs retain a qualitative ability to interpolate} even when trained on $5$k examples. Regarding memorization, it appears that DDPMs \rebuttal{trained} on $5$k examples memorize extreme examples. In \cref{sec:memorization-clip}, we show that the model avoids mere reproduction when trained on larger training sets.
\subsection{Sensitivity to the Guidance Parameters}
\label{sec:sensitivity-to-guidance}
\begin{table}
    \centering
    \begin{tabular}{|c|c|c|}
        \hline
        $|\mathcal S_\text{train}|$ & MSE to uniform $\downarrow$ & KLD to uniform $\downarrow$ \\
        \hline
        60k & 0.130 & 0.607 \\
        \hline
        30k & 0.184 & 0.835 \\
        \hline
        10k & 0.186 & 0.802 \\
        \hline
        5k & 0.279 & 1.185 \\
        \hline
    \end{tabular}
    \caption{Ablation on interpolation performance when decreasing the training set size (first column). We compute the MSE and Kullback-Leibler divergence (KLD) between the empirical histogram on $10$k samples and the uniform distribution. For both MSE and KLD, lower is better.}
    \label{tab:smile-data-size-ablation}
\end{table}
In this section, we study the sensitivity of multi-guidance to variations in $\lambda$. We generate $10$k samples for varying $\lambda$ values within the range $[3.5, 10, 30, 50, 75, 100]$ using the DDPM trained on $60$k examples, as in \cref{sec:smile-interpolation-subsec}.

Interestingly, even with $\lambda = 3.5$, multi-guidance can generate examples with mild attributes. However, we observe that higher values yield better MSE and KLD against uniform. Refer to \cref{appendix:guidance-strength-ablation-histograms} for histograms of the empirical distributions. Using $\lambda = 75$ achieves an MSE of $0.065$ and KLD of $0.188$, our best results. While the model seem able to interpolate for all values of $\lambda$, we found that using \rebuttal{the largest} values of $\lambda$ tends to generate images with artifacts more frequently than smaller values of $\lambda$. As such, in further experiments, we use $\lambda = 30$ as it achieves interpolation while introducing little artifacts.
Nonetheless, this experiment suggests that the interpolation abilities of diffusion models are robust to the choice of guidance parameter. \rebuttal{See \cref{sec:measuring-quality-and-diversity} for a discussion of quality and diversity metrics for different values of $\lambda$.}
\subsection{Interpolation on Alternative Attributes}
\label{sec:young-old-interpolation}
\begin{figure}[t] 
    \centering
    \includegraphics[width=1\linewidth]{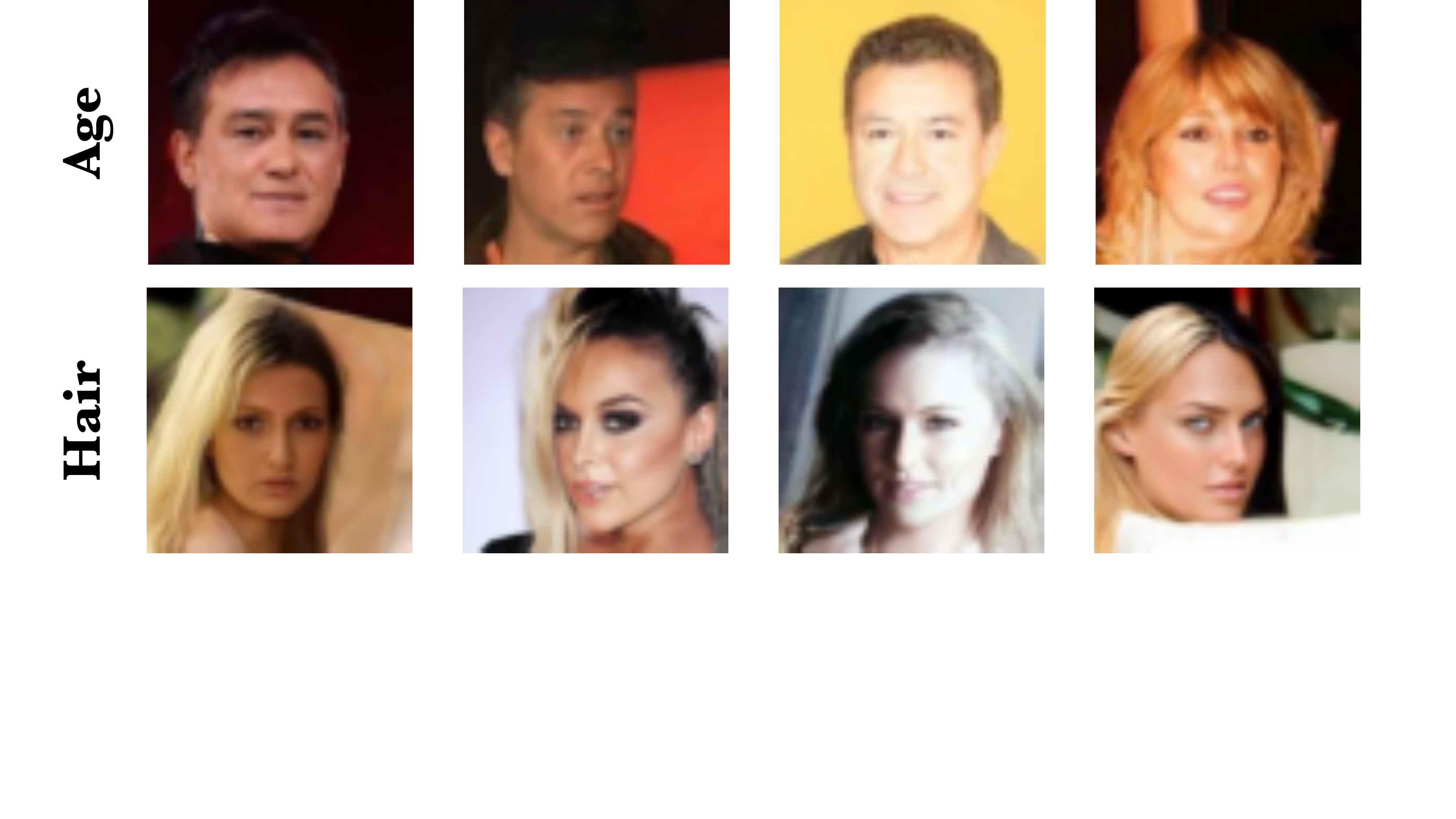}
    \vspace{-20mm}
    \caption{Samples with likelihood in $[0.49, 0.51]$ according to their respective evaluation classifier. Samples in the first row come from the "Young" interpolation experiments, while the second row interpolates according to the hair color.}
    \label{fig:young-blond-hair-pictures}
\end{figure}

To ensure our results generalize across attributes, we extend our study to interpolation based on the age and hair color on CelebA. Similar to the smile attribute, we compute the distribution of synthetic samples using an evaluation classifier\footnote{The evaluation classifiers for the "Smiling" and "Young" attributes required minimal calibration (temperature of $1.05$), while the "Blond\_Hair" and "Black\_Hair" classifiers required a temperature of $1.245$. See \citet{guo2017calibration} for details.}. Regarding hyperparameters, we reuse $\lambda = 30$ from \cref{sec:sensitivity-to-guidance}, to assess how our method generalizes to other cases without tuning. We train the DDPM as in \cref{sec:smile-interpolation-subsec}. As in previous experiments, DDPMs demonstrate interpolation abilities. Refer to \cref{fig:young-blond-hair-pictures} and \cref{subsec:additional-samples} for samples with mild attributes according to the "Young" and hair color attributes. We did not observe any instances of individuals with a clearly divided hair color, featuring half of the hair in black and the other half in blond in the training set.

\begin{figure}[t] 
    \centering
    \subfloat[]{
        \includegraphics[width=0.45\linewidth]{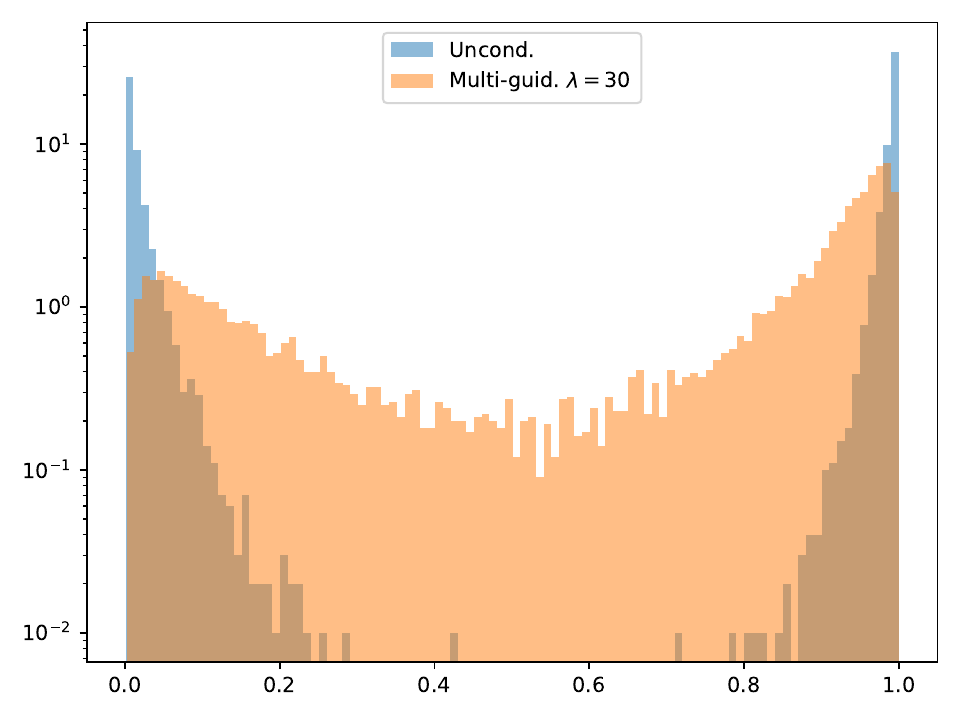}
        \label{fig:young-hist}
    }
    \hfill
    \subfloat[]{
        \includegraphics[width=0.45\linewidth]{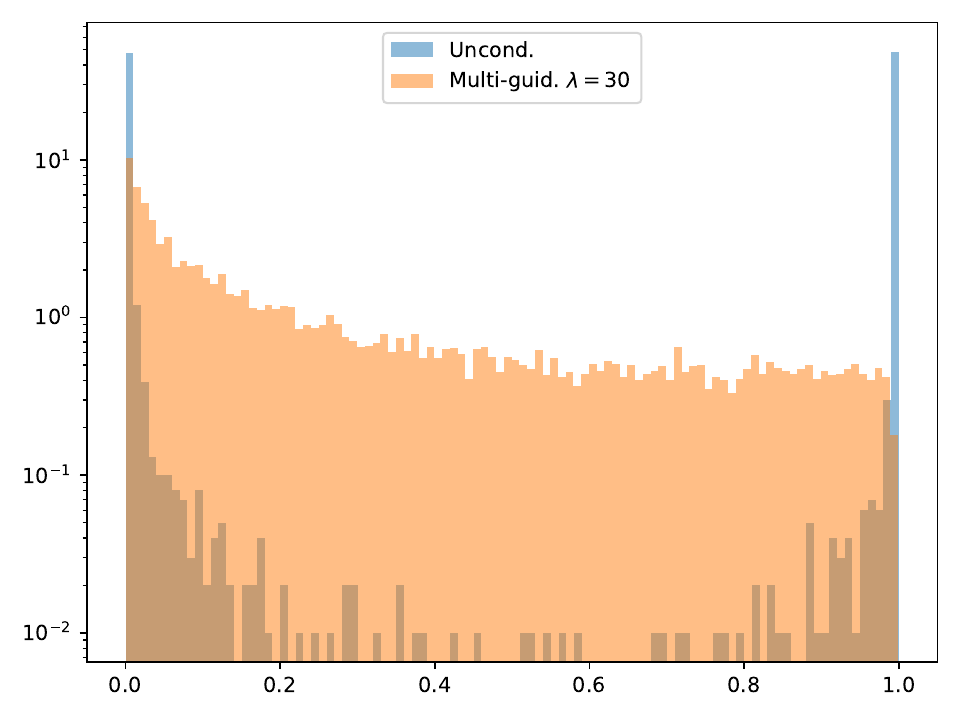}
        \label{fig:blond-hist}
    }
    \caption{Comparing train distribution and multi-guidance distribution. \textbf{Left}: "Young" attribute. \textbf{Right}: "Blond\_Hair"/"Black\_Hair" attribute. There are 28 examples from the unconditional model (out of $10$k) in $[0.2, 0.8]$ in the hair color plot. The majority of them have brown hair. See \cref{sec:young-old-interpolation} for more details.}
    \label{fig:young-hair-histograms}
\end{figure}

\paragraph{Age Attribute}
While $48\%$ of samples are labeled as "Smiling", young people are over-represented in CelebA ($78\%$). After filtering with an ensemble of classifier, we retain the $7$k oldest and the $15$k youngest examples according to the evaluation classifier. Furthermore, we over-sample the "Old" class by a factor of 2 during training. The distribution of unconditional samples is concentrated at the borders of the interval, while multi-guidance sampling generates examples in the middle of the histogram. The distribution of multi-guidance samples has an MSE of $0.146$ and a KLD of $0.656$ against uniform, showing slight differences compared to the smile interpolation experiment. Refer to \cref{fig:young-hist} for histograms of samples created with the unconditional model and multi-guidance.
\paragraph{Hair Color Attribute}
Refer to \cref{fig:blond-hist} for the empirical distribution of generated samples. Unlike for the smile and age attribute, the unconditional model trained on blond and black haired individual generates examples with mild attributes (28 samples out of $10$k within $[0.2, 0.8]$).
A manual review reveals that the majority of these samples feature faces with brown hair.
Some instances display clear blond or black hair, while others show variations like blond hair in low light or black hair in high light, which triggers the evaluation classifier.
Hence, we still believe that our filtering method avoids leakage. The distribution of multi-guidance samples has an MSE of $0.135$ and a KLD of $0.464$ against uniform.
We observed that the FaRL model was unreliable in predicting the "Blond\_Hair" and "Black\_Hair" attributes for CelebA (see \cref{appendix:farl-bad-hair-black}), hence we do not use it for filtering.
\subsection{Interpolation on a Controlled Synthetic Distribution}
\label{sec:interp-toy-dataset}
\begin{figure}[t] 
    \centering
    \subfloat[]{
        \includegraphics[width=0.18\linewidth]{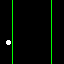}
        \includegraphics[width=0.18\linewidth]{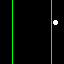}
        \includegraphics[width=0.18\linewidth]{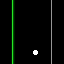}
        \label{fig:synthetic-dots}
        }
    \hfill
    \subfloat[]{
        \includegraphics[width=0.18\linewidth]{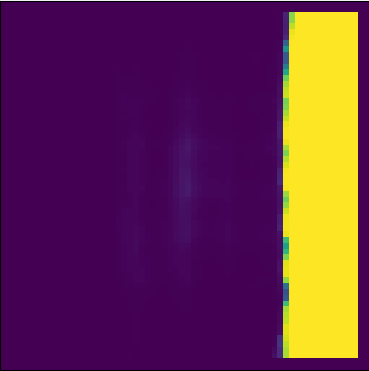}
        \includegraphics[width=0.18\linewidth]{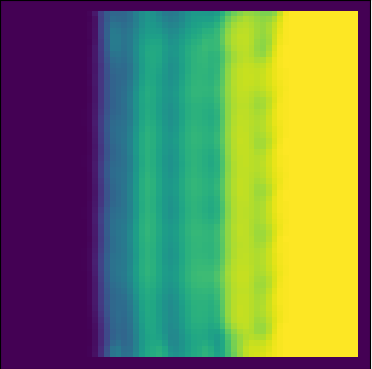}
        \label{fig:cls-reg-spectral-norm-toy}
        }
    \caption{Synthetic samples from the toy distribution. The green lines are not part of the training examples, and only serve to delimit the training regions (left/right) and unobserved region (center) for the reader. Sub-figure (b) depicts heatmaps where the color at position $(x, y)$ denotes the prediction from the guidance classifier, for images with a circle centered in $(x,y)$. Yellow colors represent high probabilities while low probabilities are shown in dark blue. The left heatmap is for a regular run while the right one is for a classifier trained with spectral normalization. We see that the transition in the unseen region (middle) is much smoother when using spectral normalization. Guidance with the left classifier yields $73\%$ of samples in the middle section while the right classifier yields slightly above $90\%$.}
        \label{fig:synthetic-data}
\end{figure}
 
We posit that interpolation occurs because, although the model does not encounter the complete $[0, 1]$ range of attributes at training, it observes sufficient variation of extreme attributes (as modeled by $\delta$ in \cref{sec:problem-statement}). Inspired by \citet{zhao2018bias}, we design a synthetic dataset of extreme examples with small variations to assess our hypothesis.

Consider $64 \times 64$ images with black background, containing a single \rebuttal{white} circle, centered in $(x, y)$, with a radius of two pixels. Let $z_i \in [0, 1]$ denote the (normalized) $x$ coordinate of the circle's center (see \cref{fig:synthetic-dots}). We construct the training set $S_\text{train}$ to contain images with $z_i \in [0, 0.2] \cup [0.8, 1.0]$ only. We train a regular DDPM and sample with multi-guidance. The classifier is trained on labels $\tilde z_i = \mathbbm{1}\left\{ z_i > 0.5 \right\}$. We train 10 classifiers and sample with multi-guidance with $\lambda = 30$ for both left and right classes. 

\rebuttal{We measure the interpolation ability of our model using the fraction of generated samples with a circle in the middle region of the image ($z_i \in [0.2, 0.8]$). We refer to the fraction as "accuracy". When sampling with multi-guidance, we achieve} an accuracy of $78\% \pm 8.7 \%$ (mean and standard deviation over 10 runs). The accuracy is computed using an edge detection algorithm implemented with OpenCV\footnote{\url{https://opencv.org}}. Note that $4.25 \% \pm 2.45\%$ of the samples feature either zero or two circles. This behavior aligns with \citet{zhao2018bias}, who made analogous observations for other classes of generative models, sometimes synthesizing samples with slightly more or fewer objects than observed in training images.

We observe that spectral normalization \citep{miyato2018spectral} on the classifier improves the smoothness of predictions in the unseen region  (see \cref{fig:cls-reg-spectral-norm-toy})  and that sampling with such a smoothed multi-guidance results in more mild samples (above 90\%). We obtain an accuracy of $75\% \pm 13.3\%$ with spectral normalization, slightly lower than regular training on average. The large variance is explained by training runs with smooth transition between classes, which achieve significantly higher accuracy, but do not occur at every run. Studying regularization methods for real-world datasets is left for future work.
\section{Interpolation with Two Factors of Interest}
\label{sec:interpolation-two-variables}
\begin{figure}[t]
    \vspace{-1mm}
    \centering
        \includegraphics[width=0.49\linewidth]{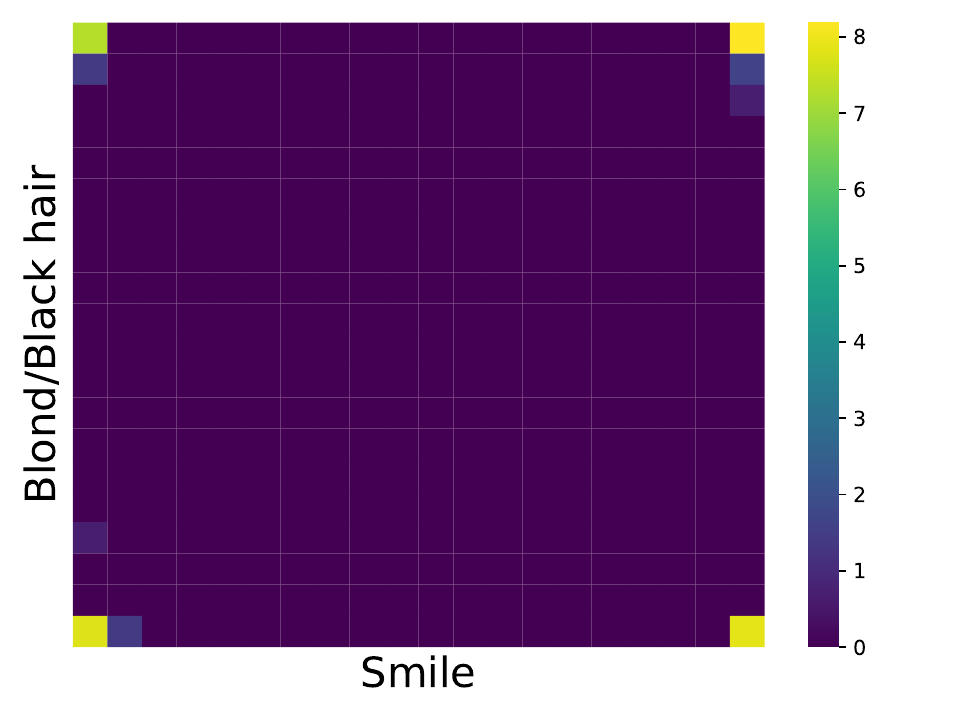}
        \includegraphics[width=0.49\linewidth]{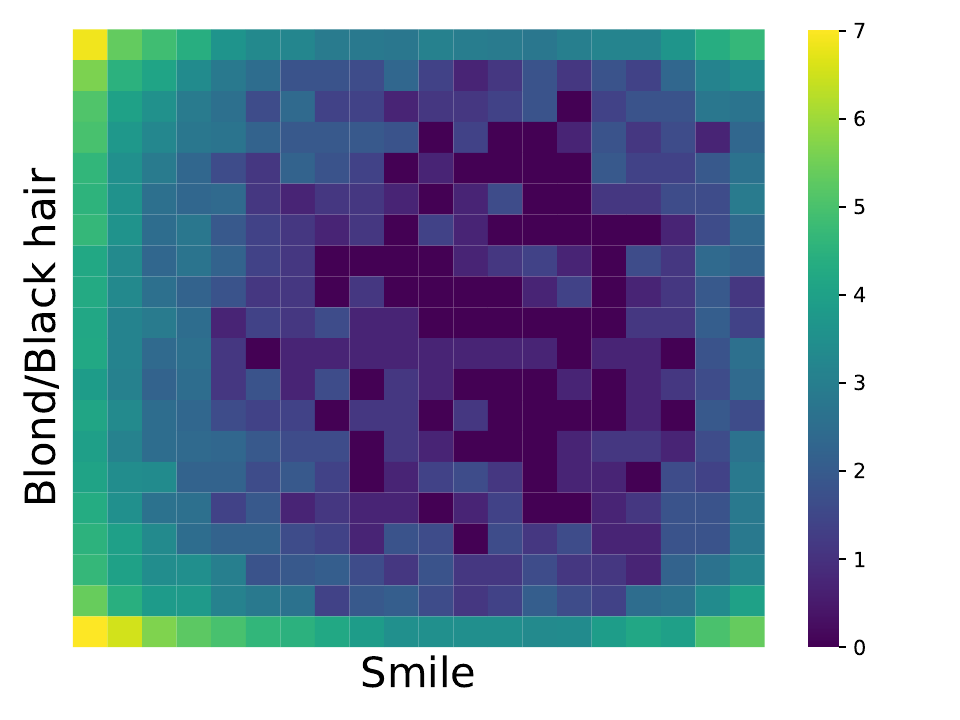}
    \vspace{-3mm}
    \caption{Log-scale heatmaps depicting results of two-variable interpolation. \textbf{(Left)}: unconditional model. \textbf{(Right)} multi-guidance with $\lambda = 30$ for all four corners. Although multi-guidance may not cover the space as comprehensively in higher dimensions, DDPM still exhibits interpolation capabilities.}
    \vspace{-3mm}
    \label{fig:2d-interpolation-heatmaps}
\end{figure}

Having explored distributions with a single variable, we extend our investigation to interpolation with two variables, specifically, the "Smiling" and "Young" attributes. Due to space constraints, a detailed discussion is deferred to \cref{sec:two-vars-interp-appendix}. While the model does not sample uniformly on the $[0, 1]^2$ square, \cref{fig:2d-interpolation-heatmaps} suggests that DDPMs can also interpolate on multiple variables. We believe that training on synthetic samples and/or regularization could bring the distribution closer to uniform.
\section{What about Extrapolation?}
\label{sec:what-about-extrapolation}
In this study, we focus on zero-shot interpolation between extreme cases, i.e. we assume access to extremal samples $\lbz,\ubz \in \mathcal{D}$ s.t. $\lbz \preceps z_{intra} \preceps \ubz$,.
On a high level, extrapolation is about generating examples with latent variables $z$ located outside the convex hull of $z_i \in \mathcal{D}$, i.e. $z_{extra}\preceps\lbz$ or $\ubz \preceps z_{extra}$ .
We hypothesize that extrapolation for DDPMs requires two key elements. First, the unconditional diffusion model should be capable to generate examples significantly different from the training distribution.
Second, we require a way to guide the diffusion model toward these distant examples.
Impressive work by \citet{zhu2023unseen} on few-shot generalization suggests that DDPMs indeed learn generic geometric priors, allowing them to sample from completely unrelated domains, if guided in that direction by using a set of reference images.
Unfortunately, in the zero shot setting it is difficult to create a guidance for latent features, as  
classifiers trained with cross-entropy tend to find the simplest model that  explains the data, a phenomenon referred to as \emph{simplicity bias} \citep{shah2020pitfalls, jacobsen2020excessive}.
One option is to directly tackle the simplicity bias, using algorithm such as \rebuttal{developed by} \citet{pagliardini2022agree}, however further constrained towards the desired latent factors. Alternatively, we believe that a classifier trained on synthetic data could guide diffusion models far from the training support. Our preliminary experiments \rebuttal{in \cref{sec:green-lips-extrapolation}} support this claim, where we successfully generated faces with green lips, despite the unconditional diffusion model lacking exposure to such instances. Nonetheless, the question of zero shot extrapolation far from the training domain is complex and left for future work.

%% file: sections/conclusion.tex
\section{Discussion}
Our work studies interpolation abilities of DDPMs. Notably, we observed that in both real-world and synthetically controlled scenarios, DDPMs are capable of interpolation, and generate more diverse data than what they were trained on. We achieve interpolation through the multi-guidance procedure (\cref{sec:multi-guidance}), and demonstrate its applicability across various datasets, including across  multiple joint latent factors. Our work contributes to the growing body of evidence suggesting that the inductive biases of diffusion models align well with real-world signals, particularly in image data, and show that a simple modification to the sampling procedure unlocks the interpolation abilities of DDPMs.

\subsection{Where are the Interpolation Capabilities coming from?}
\rebuttal{We believe that understanding the interpolation, and more generally the out-of-distribution capabilities of DMs is an exciting area of study. Our standing hypothesis regarding those capabilities is strongly influenced by the work of \citet{zhu2023unseen}. In DDIM, \citet{song2022denoising} demonstrate that any diffusion model, even an untrained one, can generate a latent representation leading to any image when sampled deterministically with DDIM. Therefore, in principle, \textbf{every diffusion model admits sampling trajectories that result in any real image}.}
\rebuttal{While most sampling paths are unlikely to be encountered in practice, \citet{zhu2023unseen} demonstrated that given a few out-of-domain images, DMs can generate out-of-domain images, even if the domain in question is significantly different from the training distribution. Crucially, the work of \citet{zhu2023unseen} defines a novel guidance procedure using a few out-of-domain images. Impressively, \citet{zhu2023unseen} successfully generate human faces with a DM trained on dog pictures. This surprising result suggests that new sampling procedures might be sufficient to unlock out-of-domain generalization. In this work, we demonstrate that DDPMs can generate samples lying outside of their training distribution, without requiring additional samples (zero-shot result). Importantly, we assume there exist underlying continuous latent variables that determine the realization of certain attributes. We further assume that only part of the latent variable support is observed during training (\cref{sec:problem-statement}).}

\subsection{Limitations and Future Work}
We defer the exploration of few-shot extrapolation, e.g. through synthetic data, to future work. Additionally, the interpolation between certain attributes is ill-defined. For instance, zero-shot interpolating between blond and black hair might produce faces with half black, half blond hair, shades of brown hair, or other results. Thus, we believe that few-shot learning could prune unwanted interpolation behavior. As for extrapolation, this could potentially be tackled with synthetic data. Moreover, we observed in \cref{sec:interp-toy-dataset} that spectral normalization improves the interpolation ability on synthetic images. Exploring this phenomenon more in depth, as well as other means of regularization is a promising venue for future work. While our approach allows sampling images with mild attributes, reliable sampling over narrow range of the latents remains elusive. Such ability could allow specifying a very precise skin tone for example. Therefore, future work may explore alternative sampling algorithms, such as in Markov Chain Monte Carlo (MCMC) methods, as studied in \citet{du2023reduce}. \rebuttal{Finally, we observed that DDPMs struggle to interpolate on the continuous range for certain datasets, such as rotation of objects in 3D.}

%% file: sections/impact_statement.tex
\section*{Impact Statement}

Our research on the interpolation abilities of diffusion models holds promise for improving fairness in machine learning. By showcasing the model's capacity to generate diverse attributes covering the range of a latent factor from a subset of examples, our work can contribute to mitigating biases in training data.
However, it's essential to recognize potential misuse. While our work is on low-dimensional images of insufficient quality for deception, the method could potentially be improved and exploited for deepfake generation in higher resolutions.
Such concerns are inherent to the development of most novel generative modeling techniques.
Importantly, our focus is not technically advancing the state of the art in image generation. Instead, we merely study the properties of diffusion models and show what is \emph{already} possible with current models.

This demonstration also has concrete implications on the use of generative models as tools in creating new content. At time of writing there are ongoing lawsuits concerning the generation of copyrighted content \citep{GenerativeAIHasa} and while the examples discussed in our citation are clearly infringing even \emph{if} they were interpolated in the manner we discuss (which we find unlikely), the ability to interpolate and potentially \emph{extrapolate} across latent factors hints at the possibility at truly novel synthesized content - which then itself would complicate discussions around the level of derivative vs. original "work" present in the generated content.

In no way should our work be cited as "proof" that any models involved in such lawsuits are not infringing.
It requires further development in dataset provenance attestation and legal frameworks to make any such claims.
Presumably, our findings will play a rule in developing these legal frameworks.
Until then, we think it is important to err on the side of caution and not train production models on any content for which the artists or copyright holders have not explicitly consented to be included as part of generative model training.

Further, while we are excited about the idea of using interpolated factors to de-bias classifiers, e.g. by developing methods for holding all factors except the interpolated one constant in order to remove spurious correlations with e.g. gender or skin color, the model used for such debiasing might itself be flawed and/or introduce new biases. This means any such use requires careful verification and ideally further work on provably selective editing via interpolation.

%% file: sections/acknowledgements.tex
\section*{Acknowledgements}
\rebuttal{We appreciate the constructive discussions with the reviewers during the rebuttal period, which contributed to the final quality of this paper. We are also grateful to the area chairs, senior area chairs, and the organizers of ICML'24 for their work. Additionally, we thank Ronan Collobert for his questions and suggestions. This work was supported by the Hasler Foundation Program: Hasler Responsible AI (project number 21043). This research was sponsored by the Army Research Office and was accomplished under Grant Number W911NF-24-1-0048. This work was supported by the Swiss National Science Foundation (SNSF) under grant number 200021\_205011.}

%% file: sections/appendix.tex
\section{Additional Background on Diffusion Models}
\label{appendix:diffusion-models}
We use the denoising diffusion probabilistic model (DDPM) formalism, introduced in \citet{ddpm} and include relevant background from \citet{improved-ddpm, dhariwal2021diffusion, du2023reduce}. For a clear long format introduction, please refer to the excellent work of \citet{luo2022understanding}.
\subsection{Forward Process}
Diffusion models approximate the data distribution $p_\text{data}$ by reversing a Markov Chain (MC), called the forward process, that transforms $p_\text{data}$ into white noise. The forward MC does not have any learnable parameters and is defined as
\begin{equation}
    p_\text{fwd}(x_0, x_1, ..., x_T) = p_0(x_0) \prod_{t = 1}^T p_t(x_t | x_{t - 1}),
    \label{eq:forward-process}
\end{equation}
where $0 \leq t \leq T$ and the variables $x_1, ..., x_T$ are corrupted with Gaussian noise. Specifically, for $t > 0$, $p_t(x_t | x_{t - 1}) := \mathcal N (x_t; \sqrt{1 - \beta_t}x_{t - 1}, \beta_t \mathbf{I})$ and $p_0(x_0) = p_\text{data}(x_0)$. The $(\beta_t)_{t = 1}^T$ coefficients are called the \textit{noise schedule} and are picked such that $p_T(x_T) \approx \mathcal N (x_T; 0, \mathbf{I})$. Using Gaussian transitions allows one to compute $x_t | x_0$ in one step. From \citet{ddpm}, defining $\alpha_t := 1 - \beta_t$, $\bar \alpha_t = \prod_{i=1}^t \alpha_i$ and $\epsilon \sim \mathcal N(0, \mathbf{I})$,  we get
\begin{equation}
    x_t = \sqrt{\bar \alpha_t}x_0 + \sqrt{1 - \bar \alpha_t} \epsilon,
    \label{eq:reparam-sample-step-t}
\end{equation}
\subsection{Backward Process}
The backward process is used to sample from $p_\text{data}$ by iteratively denoising samples from $p_T \sim \mathcal  N (0, \mathbf{I})$. Although $p_\text{fwd}(x_{t - 1} | x_t)$ depends on the entire data distribution, $p_\text{fwd}(x_{t - 1}| x_t, x_0)$ is Gaussian \citep{ddpm}. Using Bayes theorem, one finds $p_\text{fwd} (x_{t - 1} | x_t, x_0) = \mathcal N\left(x_{t - 1}; \tilde \mu(x_t, x_0), \tilde \beta_t \mathbf{I}\right)$, where
\begin{equation}
    \tilde \mu(x_t, x_0) := \frac{\sqrt{\bar \alpha_t}}{1 - \bar \alpha_t} \left(x_0 + (1 - \bar \alpha_{t - 1})x_t \right), %
\label{eq:mu-posterior}
\end{equation}
\begin{equation}
    \tilde \beta_t := \frac{1 - \bar \alpha_{t - 1}}{1 - \bar \alpha_t} \beta_t.
\end{equation}
DDPMs approximate $p_\text{fwd} (x_{t - 1} | x_t, x_0)$ with a distribution  $p_\theta (x_{t - 1} | x_t)$, implemented with a denoising neural network. Formally, $p_\theta (x_{t - 1} | x_t) \sim \mathcal N \left(x_{t - 1} ; \mu_\theta (x_t, t), \Sigma_\star (x_t, t)\right)$. \citet{ddpm} use a fixed covariance, being either $\Sigma_\star (x_t, t) = \beta_t \mathbf{I}$ or $\Sigma_\star (x_t, t) = \tilde \beta_t \mathbf{I}$. The neural network typically computes $\mu_\theta (x_t, t)$ in one of three ways. First, output $\mu_\theta (x_t, t)$ directly. Second, approximate $x_\theta (x_t, t) \approx x_0$ and compute $\mu_\theta (x_t, t)$ by replacing $x_0$ in \cref{eq:mu-posterior}. Finally, predict the noise to remove from $x_t$ to obtain $x_0$, denoted by $\epsilon_\theta (x_t, t)$. Manipulating \cref{eq:reparam-sample-step-t} and \cref{eq:mu-posterior} yields 
\begin{equation}
    \mu_\theta(x_t, t) = \frac{1}{\sqrt{\alpha_t}}\left(x_t - \frac{\beta_t}{\sqrt{1 - \bar \alpha_t}}\epsilon_\theta(x_t, t) \right).
\end{equation}
Note that \citet{ddpm} obtain their best results when modeling $\epsilon_\theta (x_t, t)$. While \citet{ddpm} keep $\Sigma_\star (x_t, t)$ fixed, \citet{improved-ddpm} use a learnable $\Sigma_\star (x_t, t) = \Sigma_\theta (x_t, t)$, computed as 
\begin{equation}
    \Sigma_\theta (x_t, t) = \exp \left( v \log \beta_t + (1 - v) \log \tilde \beta_t\right),
    \label{eq:improved-ddpm-covariance}
\end{equation}
where $v$ comes from a neural network.
\subsection{Learning Objective} 
One can directly optimize the variational bound $L_\text{vlb}$, which writes as
\begin{equation}
    \begin{aligned}
        &L_\text{vlb} &:= & ~ L_0 + L_1 + ... + L_{T - 1} + L_{T} \\
        &L_0 &:= & ~ \mathbb E_{p_\text{fwd}(x_1 | x_0)} \left[ - \log p_\theta (x_0 | x_1) \right] \\
        &L_{t - 1} &:= & ~ D_\text{KL}\left(p(x_{t - 1}| x_t, x_0) ~ \| ~ p_\theta (x_{t - 1} | x_t)\right) \\ %
        &L_T &:= & ~ D_\text{KL}\left( p(x_T | x_0) ~ \| ~ p_T (x_T) \right),
    \end{aligned}
    \label{eq:vlb}
\end{equation}
where the Kullback-Leibler divergence (KLD) $D_\text{KL}$ between Gaussians admits a simple expression \citep{kl-gaussians-duchi}. Choosing a noise schedule $(\beta_t)_{t = 1}^T$ such that $x_T$ is white noise makes $L_T$ independent of any learnable parameters. While optimizing $L_\text{vlb}$ is possible, \citet{ddpm} obtained their best results in terms of \textit{Fréchet Inception Distance} (FID) \citep{ttur-gan-fid} using a fixed covariance $\Sigma_\star (x_t, t)$ and the objective $L_\text{simple} := \mathbb E_{t, x_0, \epsilon} \left[ \| \epsilon - \epsilon_\theta (x_t, t) \|^2 \right]$, where $t$ is picked uniformly at random in $1, ..., T$, $x_0 \sim p_\text{data}$ and $\epsilon \sim \mathcal N (0, I)$ is the noise used to compute $x_t$ through \cref{eq:reparam-sample-step-t}. As shown in \citet{ddpm}, $L_\text{vlb}$ is equivalent to $L_\text{simple}$ up to a reweighing of the $L_t$ terms. While competitive in terms of FID, training DDPMs with $L_\text{simple}$ achieves lower likelihood than likelihood-based models such as \citet{oord2016conditional}. \citet{improved-ddpm} achieved competitive likelihood while retaining the FID of \citet{ddpm} with three improvements. First, a novel noise schedule $(\beta_t)_{t = 1}^T$, especially suited to $64 \times 64$ and $32 \times 32$ images, defined as 
\begin{equation}
    f(t) = \cos^2\left(\frac{1 + \frac{t}{T} + s}{1 + s} \cdot \frac{\pi}{2} \right), ~ \bar{\alpha}_t = \frac{f(t)}{f(0)}.
\label{eq:cosine-noise-schedule}
\end{equation}
Second, using the learned covariance $\Sigma_\theta(x_t, t)$ from \cref{eq:improved-ddpm-covariance}, and third, optimizing $L_\text{hybrid} := L_\text{simple} + \lambda L_\text{vlb}$. Importantly, they used a stop-gradient on $\mu_\theta (x_t, t)$ in $L_\text{vlb}$ so that it drives the learning of $\Sigma_\theta(x_t, t)$ only. \citet{improved-ddpm} used $\lambda = 0.001$ so that $L_\text{vlb}$ does not overwhelm $L_\text{simple}$.
\section{Additional Background on the Kullback-Leibler Divergence}
\label{sec:kld-explanation}

In generative modelling, one wants to approximate an empirical distribution $p_\text{data}$ using a learned distribution $p_\theta$. To do so, a common objective is the Kullback-Leibler divergence (KLD), defined in the discrete case as

\begin{equation}
    D_\text{KL}(p || q) := \sum_x p(x) \log \frac{p(x)}{q(x)}.
\end{equation}

The Kullback-Leibler divergence is relevant for generative modeling because it is minimized when $p$ and $q$ are equal. Note that the KLD is not symmetric with respect to $p$ and $q$. Hence, when approximating $p_\text{data}$, it remains important to realize whether we should minimize $D_{KL}(p_\text{data} || p_\theta)$, called \textbf{forward} KLD or $D_{KL}(p_\theta || p_\text{data})$, called \textbf{reverse} KLD. To understand the difference between the two, it is good to think of extreme examples. Those examples are inspired from \href{https://agustinus.kristia.de/techblog/2016/12/21/forward-reverse-kl/}{this useful resource}.

\paragraph{Forward KLD}
Imagine that sample $x_0$ captures a large portion of the total mass of $p_\text{data}$, e.g. if $x_0$ comes from one of the modes of the distribution. At the same time, imagine that the learned distribution $p_\theta$ does not capture this part of the distribution, hence $p_\theta(x_0) \approx 0$. Then, the term associated with $x_0$ in the KLD, i.e. $p_\text{data}(x_0) \log \frac{p_\text{data}(x_0)}{p_\theta(x_0)}$ will be large. As such, the forward KLD is called \textbf{zero-avoiding}, because to minimize the forward KLD, $p_\theta$ should assign positive mass to all $x$ such that $p_\text{data}(x) > 0$.

\paragraph{Reverse KLD}
Imagine that we want to minimize $D_{KL}(p_\theta || p_\text{data})$. Assume that $p_\text{data}(x_0) \approx 0$. Recall that $x_0$ only appears in the term $p_\theta(x_0) \log \frac{p_\theta(x_0)}{p_\text{data}(x_0)}$ of the KLD. If the learned distribution assigns weight to $x_0$, then the term will be large. Therefore, the reverse KLD is said to be \textbf{zero-forcing}, i.e. the learned distribution is discouraged to put any mass on unlikely samples.

\paragraph{KLD used by Diffusion Models}
The KL formulation used in DDPMs \cite{ddpm, improved-ddpm} is the forward KL divergence, i.e. assuming the model converges and has enough capacity, it should cover all modes of the data distribution.

\paragraph{KLD used in Evaluations}
Since we train DDPMs with the forward KL divergence, we also compare the empirical distribution of histograms using the forward KLD: $D_{KL}(p_\text{hist} || p_\mathcal{U})$, where $p_\text{hist}$ is the empirical distribution with respect to samples and $p_\mathcal{U}$ is uniform with the same support as $p_\text{hist}$.
\section{Details on Compute and Training}
\label{sec:compute-hparams}
\paragraph{Compute Costs}
Training the unconditional diffusion model on $64 \times 64$ images required using 2 GPUs and 18GiB of memory on each when using FP16. Training for $250$k steps ranged from 19 to 21 hours on A100 40GiB or RTX4090 respectively. In parallel, training the classifier for $150$k steps took 8 hours on a single GPU in FP32 (less stable) and required 12 GiB of memory. In contrast, training the unconditional model on $128 \times 128$ images required 41GiB of memory and took 34 hours, while the classifier took $13$ hours to train on $128 \times 128$ images. Finally, multi-guidance does not introduce additional cost at sampling versus regular classifier guidance, since the score with respect to all classes can be computed in a single forward/backward pass. 
\paragraph{Hyperparameters}
In this work, we train DDPMs with classifier guidance following best practices from \citet{improved-ddpm}. We train the diffusion model for $250$k steps with learned denoising process variance, a learning rate of $1e-4$, no weight decay, an EMA rate of $0.9999$, \num{4000} diffusion steps, and the cosine noise schedule presented in \cref{eq:cosine-noise-schedule}, well-suited for $64 \times 64$ images. Our model also predicts the variance of the reverse process and optimizes the objective $L_\text{hybrid}$. We use time embedding vectors of dimension 128 and 3 residual blocks per resolution, with attention at resolution \num{32}, \num{16}, and \num{8}. We train the guidance classifier for $150$k steps, with a learning rate of $3e-4$ and no weight decay. Both models are trained with the Adam optimizer \citep{kingma2017adam}. Further details on these parameters can be found in \citet{improved-ddpm}. Our GitHub repository\footnote{The URL will be provided upon de-anonymization of this work} contains the code and configuration to reproduce our results. We sample new images using \num{250} steps instead of \num{4000}, as it induces minimal change in FID.
\section{Evaluation and Data Processing Pipelines}
\subsection{Extracting an Extreme "Smiling" Subset from CelebA}
\label{appendix:extract-extreme-dataset}
Despite being labeled with binary attributes, the samples in real-world datasets such as CelebA \citep{liu2015celeba} cover a wide range of variations. For instance, the CelebA dataset contains clearly as well as mildly smiling faces, in the sense of \cref{sec:problem-statement}. In this section, we present our filtering method for the "Smiling" attribute.

Our approach involves two steps and is inspired by \citet{lakshminarayanan2017simple, fang2023data}. Initially, we train a smile classifier on the entire CelebA, denoted as $\mathcal D = \{(x_i, y_i): x_i \in \mathbb R^d, y_i \in \{0, 1\}, 1 \leq i \leq N \}$, where $N \approx 163$k. We filter the dataset to keep examples where the classifier is confident and agrees with the CelebA labels. That is, when $p_\theta( y_i | x_i) > \tau_1$. The resulting dataset is denoted as $\mathcal D_\text{filter}$. In the second step, we randomly extract 5 subsets from $\mathcal D_\text{filter}$ and train 5 classifiers, denoted as $p_{\theta_k}$, for $1 \leq k \leq 5$. The final dataset, $\mathcal D^\star \subseteq \mathcal D$, is made of examples where the 5 classifiers are confident and agree with the original labels, i.e., $p_{\theta_k}(y_i | x_i) > \tau_2$. The thresholds $\tau_1$ and $\tau_2$ are chosen such that both the original classifier and the ensemble of $5$ achieve a precision of \num{1} on a held-out, manually labeled dataset of \num{200} examples. We manually construct this held-out set with samples where people are either clearly smiling (e.g. large grin with clearly visible teeth) or clearly non-smiling. We explicitly discard samples where there we are in doubt whether the person is smiling or not. Additionally, we label the images three times to minimize the risk of mistake.

After training all the classifiers, we visualized examples from the validation set where the ensemble \emph{confidently} disagrees with the CelebA labels. There were no such images labeled as smiling, yet around \num{7000} images labeled as non-smiling were misclassified. Upon manually inspecting \num{500} of these, we agreed with the classifier's prediction more than the original labels. This comforts us in the soundness of our method. Subsequently, we examined \num{500} randomly chosen samples from $\mathcal D^\star \subseteq \mathcal D$ for each class, and concluded that all of them correspond to extreme examples, clearly smiling or non-smiling.
\subsection{FaRL Failures on Hair Color}
\label{appendix:farl-bad-hair-black}
\begin{figure}[t] 
    \centering
    \includegraphics[width=1\linewidth]{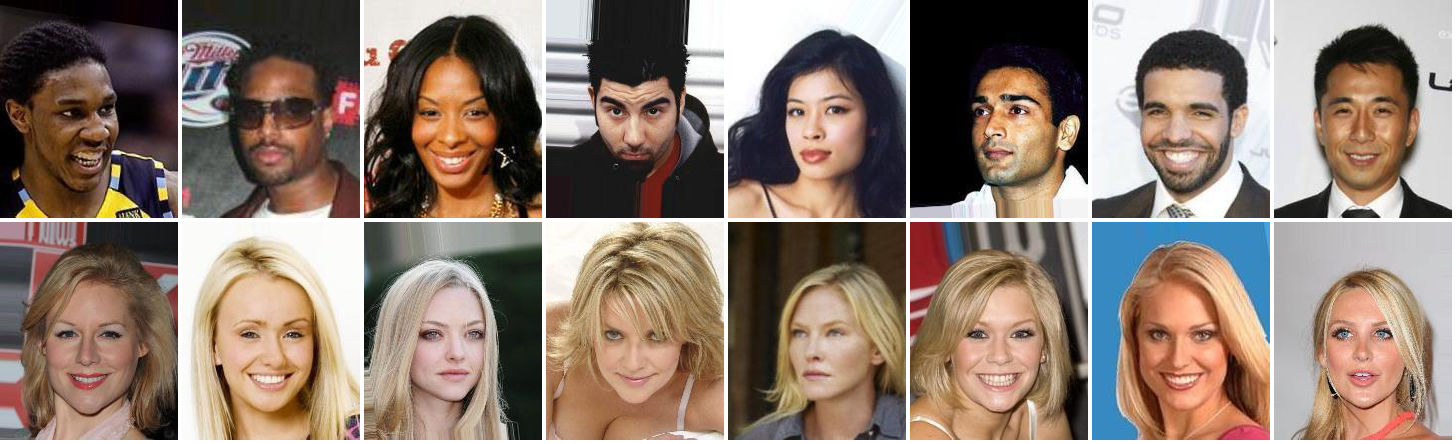}
    \caption{Examples of images where FaRL fails blatantly. The first row depicts individuals with black hair, while the second row shows some with blond hair. For both rows, FaRL assigns a hair color likelihood $< 0.01$.}
    \label{fig:farl-black-blond-hair-appendix}
\end{figure}
The FaRL model by \citet{zheng2022general}, as stated by the authors, is trained to "learn a universal facial representation that boosts all face analysis tasks." While the model was found to be a strong backbone for age estimation by \citet{paplham2023reflect}, we observed that FaRL is less reliable than a classifier trained on CelebA for detecting the smile and hair color attributes. When discriminating between the blond and black attributes on full sized CelebA images ($218 \times 178$), FaRL achieves an accuracy of $0.89$ on the validation set. We strictly followed the instructions on the official GitHub repo. On the other hand, our simple EfficientNet classifier trained on downscaled $64 \times 64$ images achieves an accuracy of $0.98$. We observed similar results for the smile attribute. Thus, we only use FaRL for age estimation. \cref{fig:farl-black-blond-hair-appendix} display images where FaRL predicts a probability above $0.999$ for the incorrect class. We do not believe those examples are particularly challenging. 
\subsection{Rough Estimate of Errors of The Filtering Method}
\label{sec:napkin-estimate-filtering-error}
\citet{recht2019imagenet} studies on ImageNet \citep{russakovsky2015imagenet} that when transferring from the train distribution to an unseen test set, an accuracy drop of 3\% to 15\% is expected. Note that CelebA is a simpler dataset than ImageNet since it contains only faces while ImageNet contains diverse classes going from specific breed of dogs to drumsticks. Let's assume the worst-case, i.e. that each classifier in the ensemble is accurate only 85\% of the time only. A sample with mild attributes remains in $S^\star$ only if each of the five models confidently classifies it as having clear attributes. Assuming reasonable independence among classifiers, given that models are not trained on the same data, one can approximate the error probability for a single sample as $(\nicefrac{15}{100})^5$. We approximate the probability that at least $k$ samples are incorrectly included in the dataset, $P(X \geq k)$, using the Chernoff bound \citep{Molloy2002}. This approximation is reasonable with our parameters ($p=(\nicefrac{15}{100})^5$, $n=$ $60$k, number of samples kept, $k>np$, number of errors). We find that $P(X \geq 10) \approx 0.089$, $P(X \geq 15) \approx  5e-4$ and $P(X \geq 20) \approx 7.2e-7$. 
\section{Extrapolation on the Lips Color Attribute}
\label{sec:green-lips-extrapolation}
While investigating interpolation, we conducted preliminary experiments on extrapolation using synthetic data for training the guidance classifier. Specifically, we generated synthetic data utilizing dlib\footnote{\url{http://dlib.net}} to detect lips in CelebA pictures and overlay a green polygon on them. Intriguingly, using the "green lips" classifier enables the DDPM to generate images with green lips, as depicted in \cref{fig:green-lips}. Although the green patch may not always be perfectly centered on the lips, the unconditional diffusion model never encountered faces with green lips, hence the process is solely driven by the classifier. 
\section{Additional Material on Interpolation on Two Variables}
\label{sec:two-vars-interp-appendix}
We explore interpolation involving two variables. To create the training dataset, we retain extremes samples for both smile and hair color attributes. As such, we simply retain the intersection of extreme examples from experiments on smile and hair color, resulting in approximately $11,000$ examples. The heatmap visualizations in \cref{fig:2d-interpolation-heatmaps} illustrate samples categorized by hair color and smile evaluation classifiers. We train the guidance classifier on 4 classes, corresponding to the corners of the heatmap. With a guidance parameter set to $\lambda = 30$ across all classes, we observe interpolation abilities of the model. %

\section{Additional Samples and Histograms}
\label{sec:additional-samples-histograms}
In this section, we present additional samples and histograms that could not fit in the main body for space reasons.
\subsection{Ablation on the Training Data Size}
\label{appendix:smile-data-size-histograms}
\begin{figure}[t]
    \centering
    \subfloat[]{
        \includegraphics[width=0.3\linewidth]{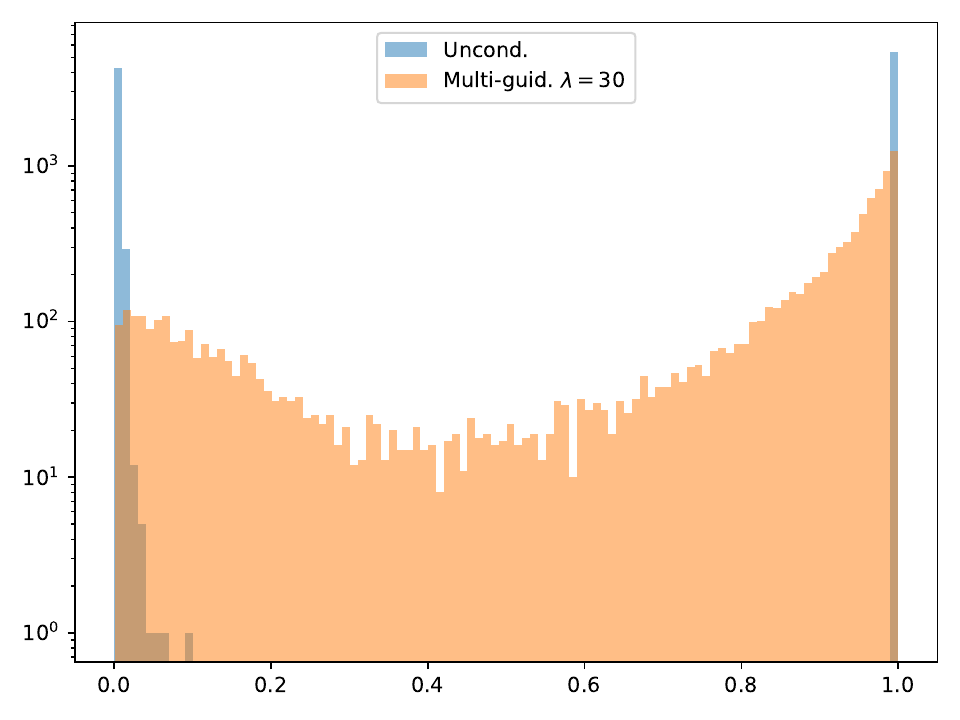}
    }
    \subfloat[]{
        \includegraphics[width=0.3\linewidth]{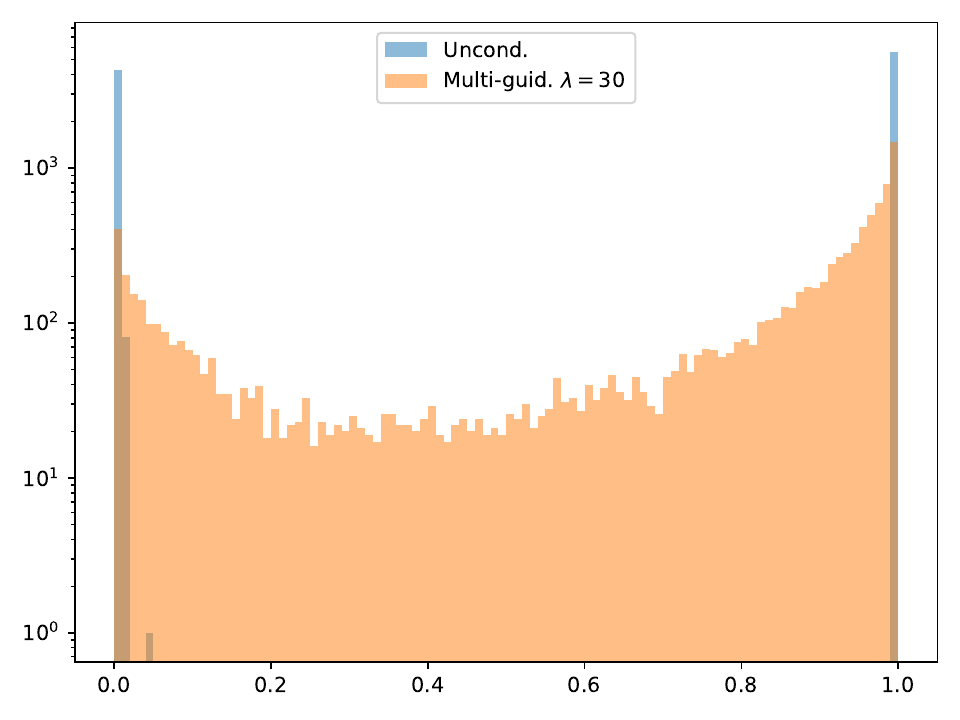}
    }
    \subfloat[]{
        \includegraphics[width=0.3\linewidth]{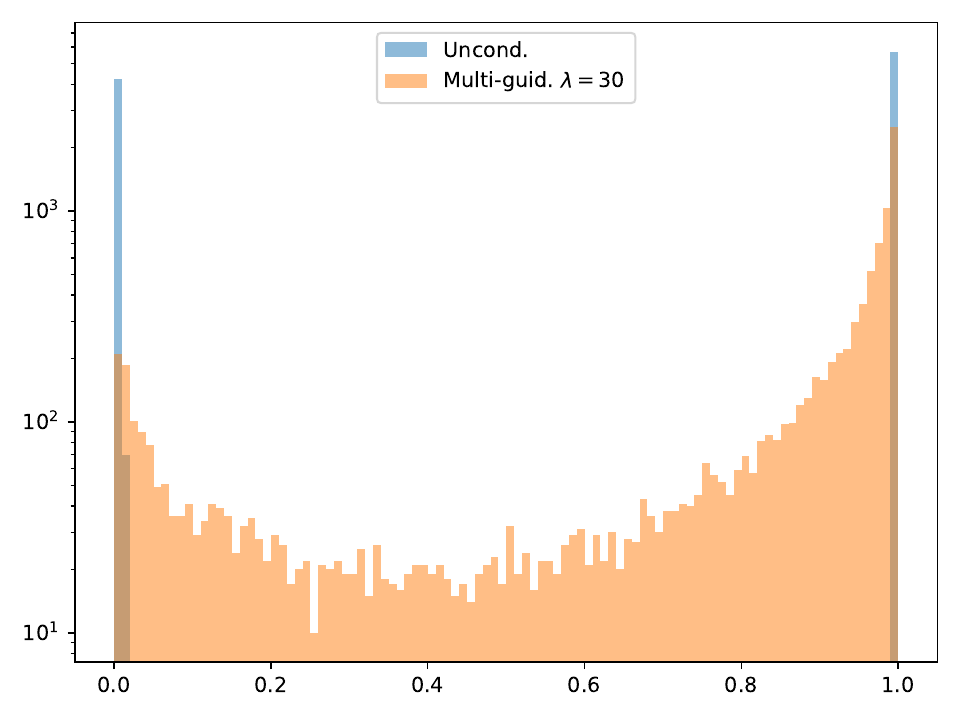}
    }
    \caption{Multi-guidance performance when reducing the training data size. \textbf{(a)}: With 30k training samples. \textbf{(b)}: With 10k training samples. \textbf{(c)}: With 5k training samples. While using 30k or 10k maintains similar performance, the interpolation capability starts to degrade with 5k, even though DDPM retains its ability to interpolate.}
    \label{fig:smile-data-size-ablation}
\end{figure}
We present histograms of the empirical distributions according to the evaluation classifier on the smile attribute in \cref{fig:smile-data-size-ablation}. As we saw in \cref{sec:how-little-data-can-we-use}, the KLD and MSE is similar when training on $30$k or $10$k examples, and deteriorates substantially when training on \num{5000} samples only. Naturally, the best performance in terms of KLD and MSE is achieved with $60$k examples, as the DDPM is exposed to more diversity.
\subsection{Ablation on the Classifier Guidance Strength Parameter}
\label{appendix:guidance-strength-ablation-histograms}
\begin{figure}[t]
    \centering
    \includegraphics[width=0.3\linewidth]{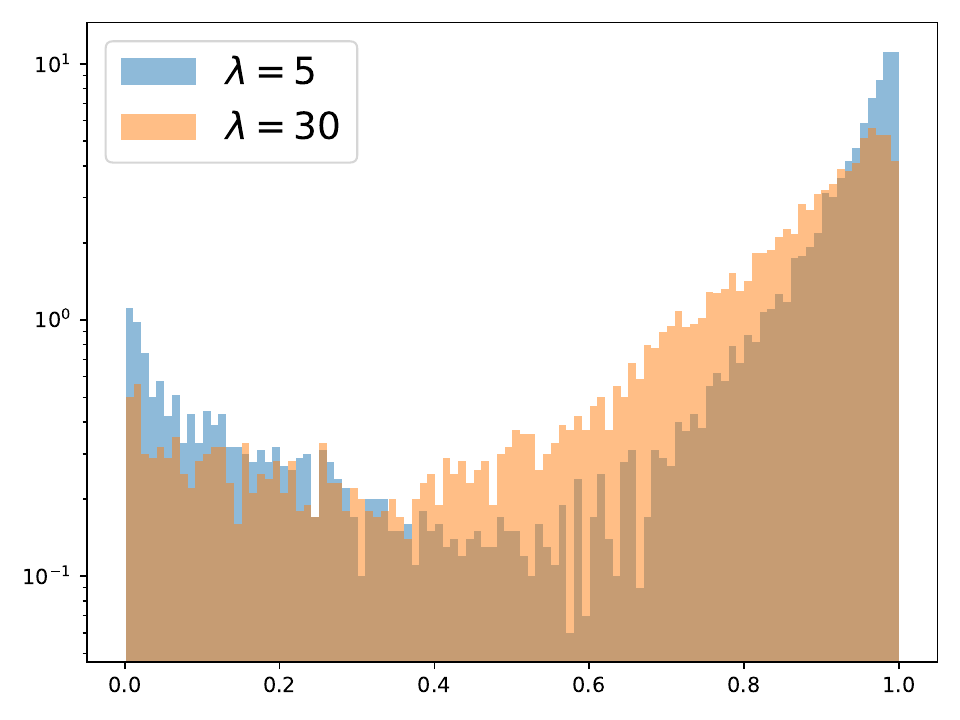}
    \includegraphics[width=0.3\linewidth]{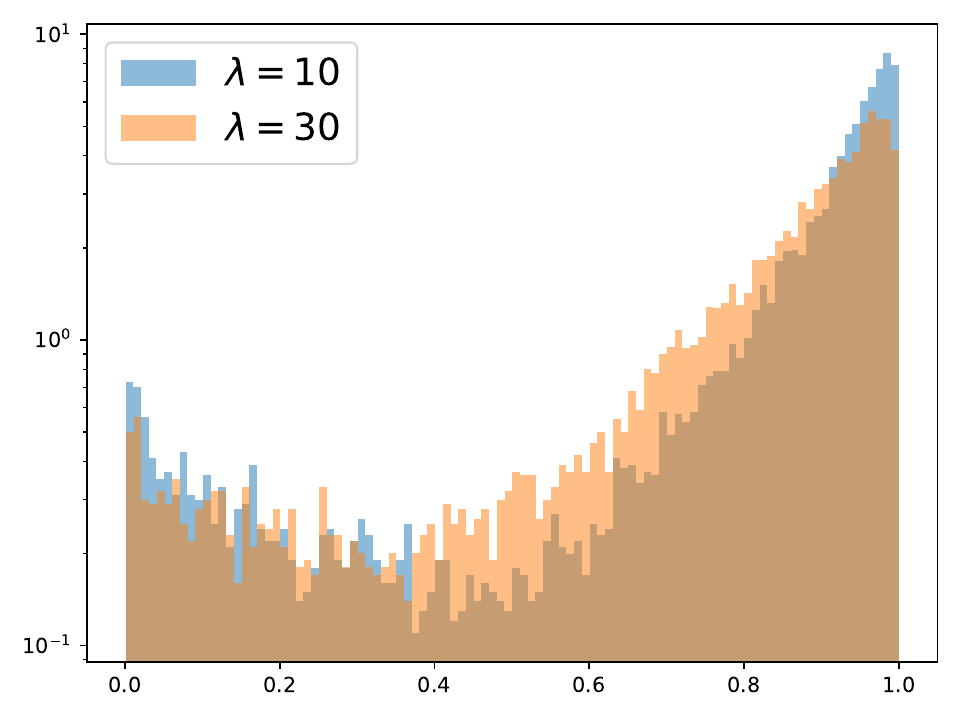}
    \includegraphics[width=0.3\linewidth]{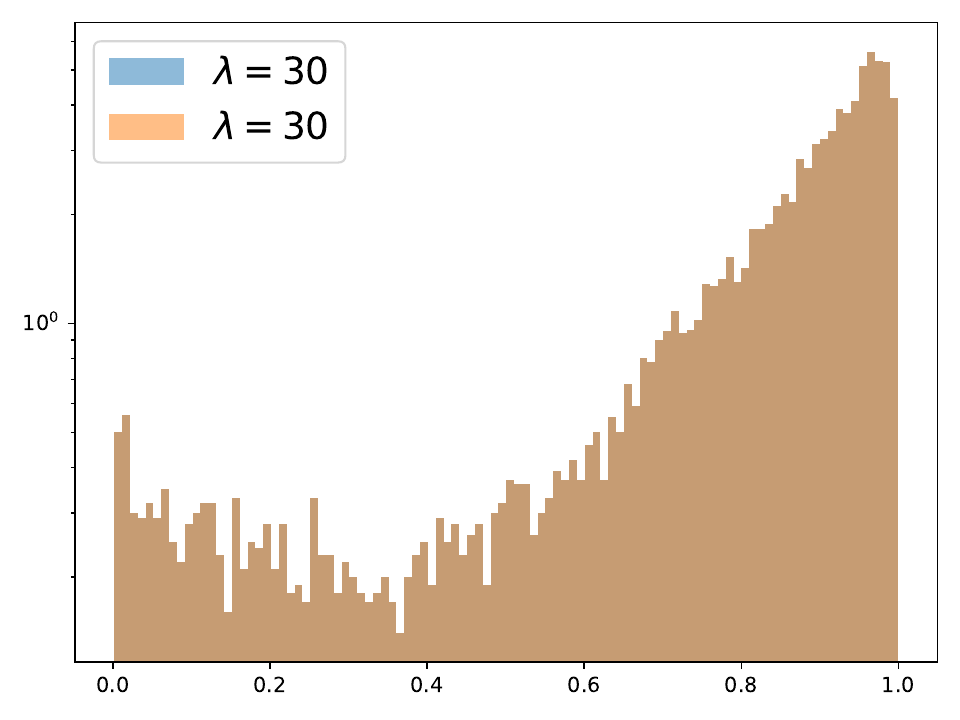}
    \includegraphics[width=0.3\linewidth]{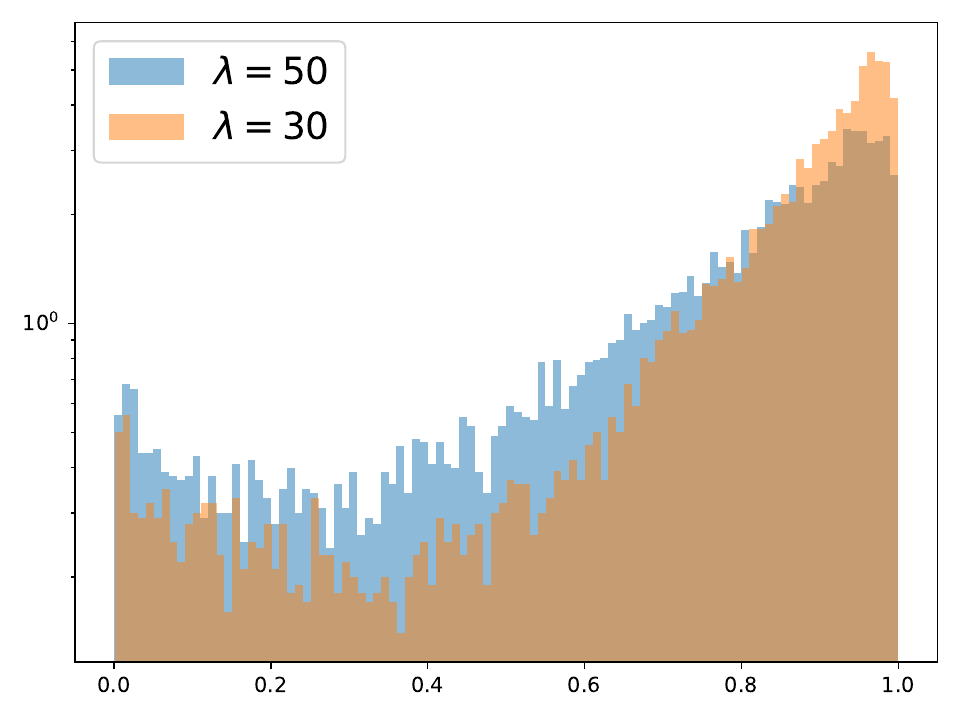}
    \includegraphics[width=0.3\linewidth]{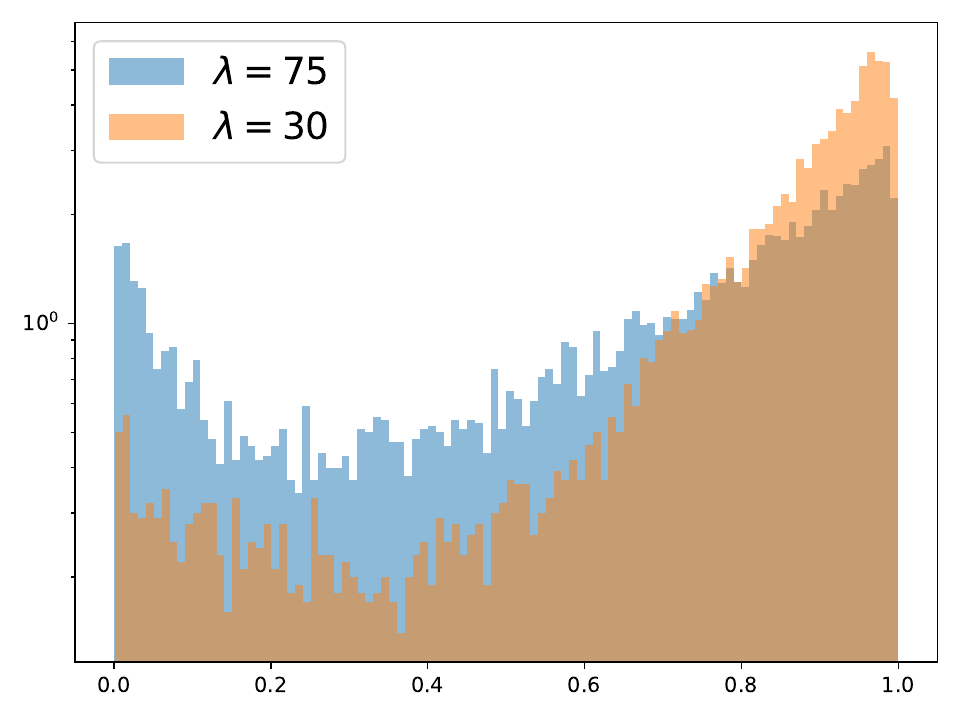}
    \includegraphics[width=0.3\linewidth]{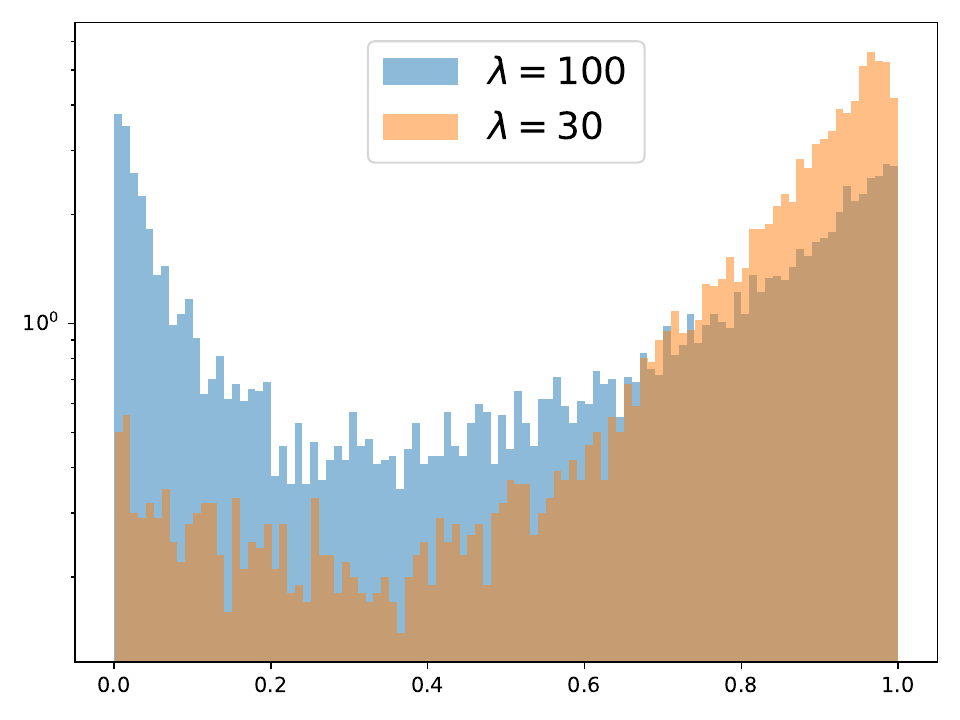}
    \caption{Histograms of the smile attribute when varying the guidance parameter $\lambda$.}
    \label{fig:hist-guid-ablation}
\end{figure}

\begin{figure}[t]
    \centering
    \includegraphics[width=0.5\linewidth]{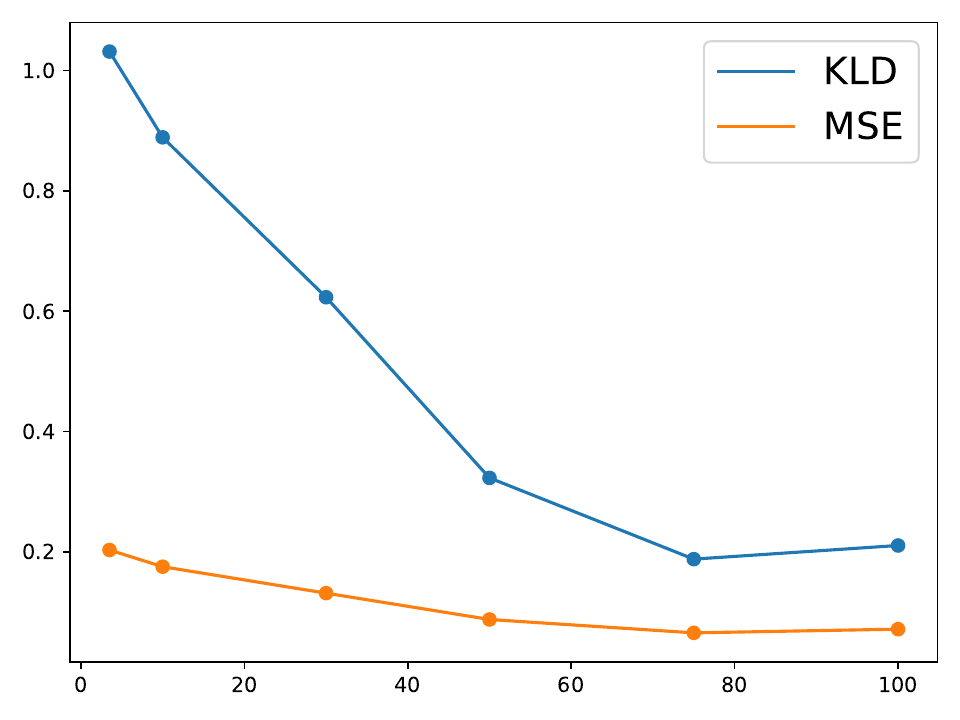}
    \caption{KLD and MSE between the empirical distribution (histograms in \cref{fig:hist-guid-ablation}) and the uniform distribution. The best performance is achieved for $\lambda = 75$.}
    \label{fig:mse-kld-guid-ablation}
\end{figure}

In \cref{fig:mse-kld-guid-ablation}, we observe that increasing the guidance strength almost monotonically improve the interpolation performance according to the MSE and KLD. In \cref{fig:hist-guid-ablation}, we display the empirical distribution of samples according to the smile attribute, using models trained on $60$k examples.
[t]
\subsection{Additional Samples}
\label{subsec:additional-samples}
\begin{figure}[t] 
    \centering
    \subfloat[]{
        \includegraphics[width=0.45\linewidth]{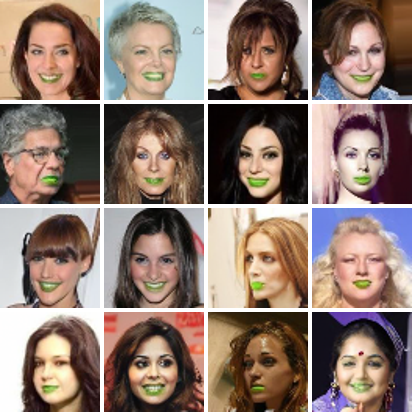}
        \label{fig:green-lips}
    }
    \hfill
    \subfloat[]{
        \includegraphics[width=0.45\linewidth]{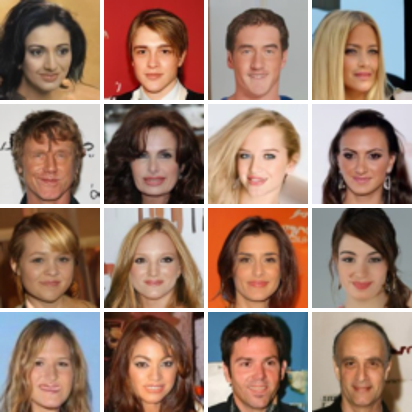}
        \label{fig:mild-smile}
    }
    \vfill
    \subfloat[]{
        \includegraphics[width=0.45\linewidth]{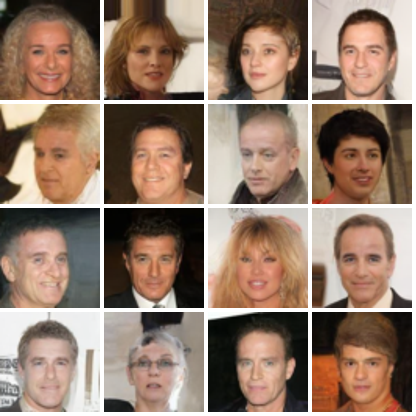}
        \label{fig:mild-age}
    }
    \hfill
    \subfloat[]{
        \includegraphics[width=0.45\linewidth]{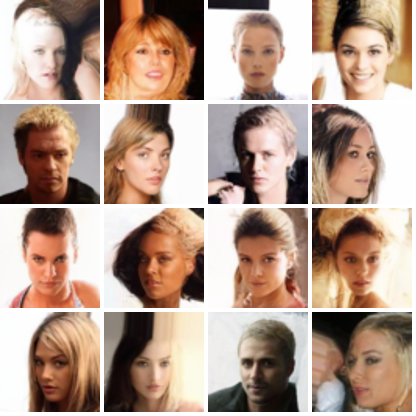}
        \label{fig:mild-hair}
    }
    \caption{Additional \emph{synthesized} samples. \textbf{(a)}: Extrapolation with classifier trained on synthetic data. \textbf{(b)}: Interpolation on smile attribute. \textbf{(c)}: Interpolation on age attribute. \textbf{(d)}: Interpolation on hair color.}
    \label{fig:additional-samples}
\end{figure}

In this section, we present additional samples that could not fit in the main body. Specifically, in \cref{fig:green-lips}, we showcase samples with \textbf{green lips}, from our preliminary extrapolation experiment. In \cref{fig:mild-smile}, we display examples with \textbf{mild smiles}. In \cref{fig:mild-age}, we present interpolated samples between \textbf{young and old} individuals. Finally, in \cref{fig:mild-hair} we showcase examples when interpolating between the \textbf{blond and black hair} colors.
\section{Checking for Memorization}
\label{sec:memorization-clip}
\begin{figure}[t]
    \centering
    \includegraphics[width=1\linewidth]{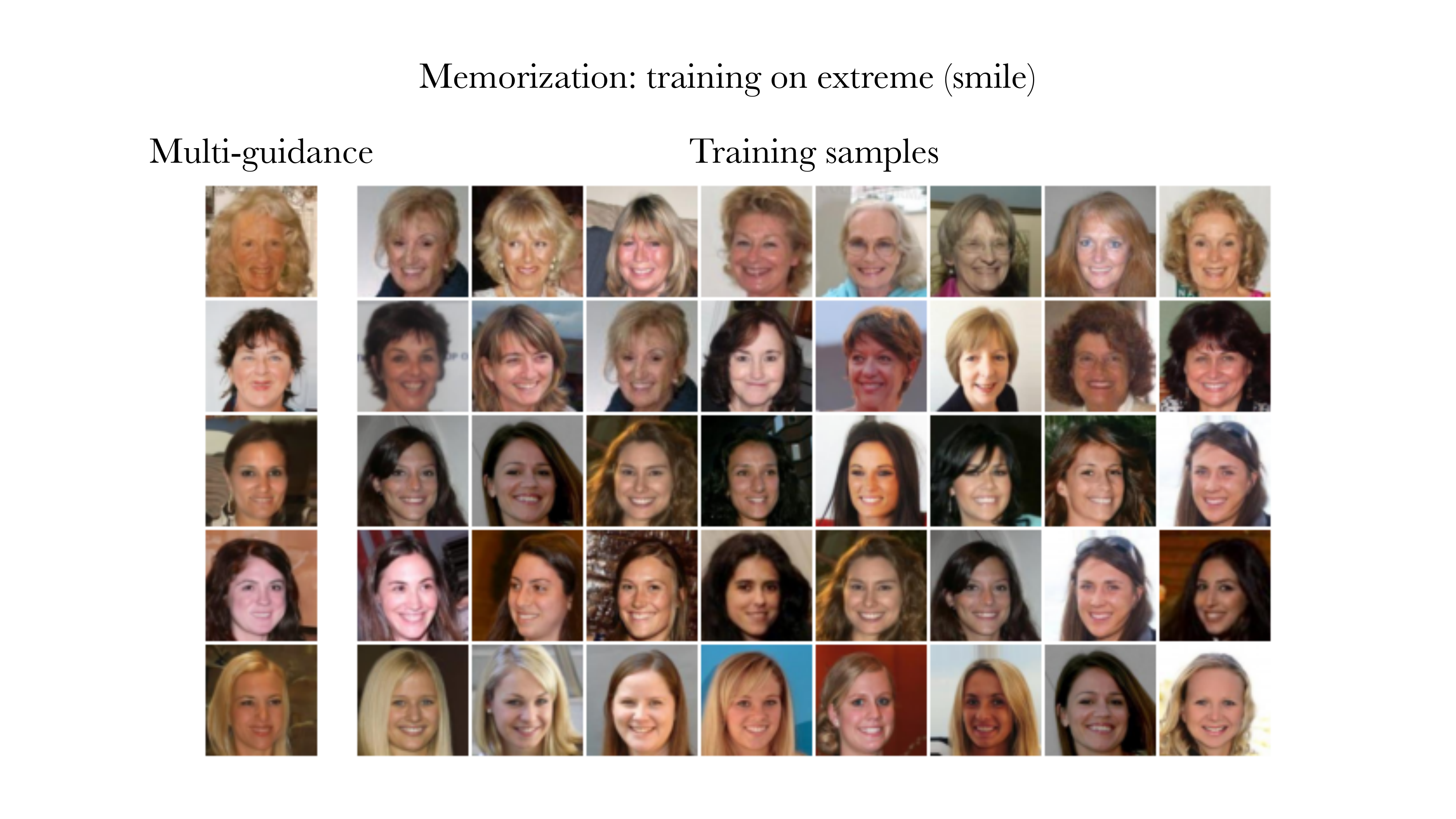}
    \caption{Checking for memorization when training on $60$k extreme examples from CelebA. The generated samples are on the left, while the right side showcases the nearest neighbors according to CLIP embeddings}
    \label{fig:smile-60k-memorization}
\end{figure}

\begin{figure}[t]
    \centering
    \includegraphics[width=1\linewidth]{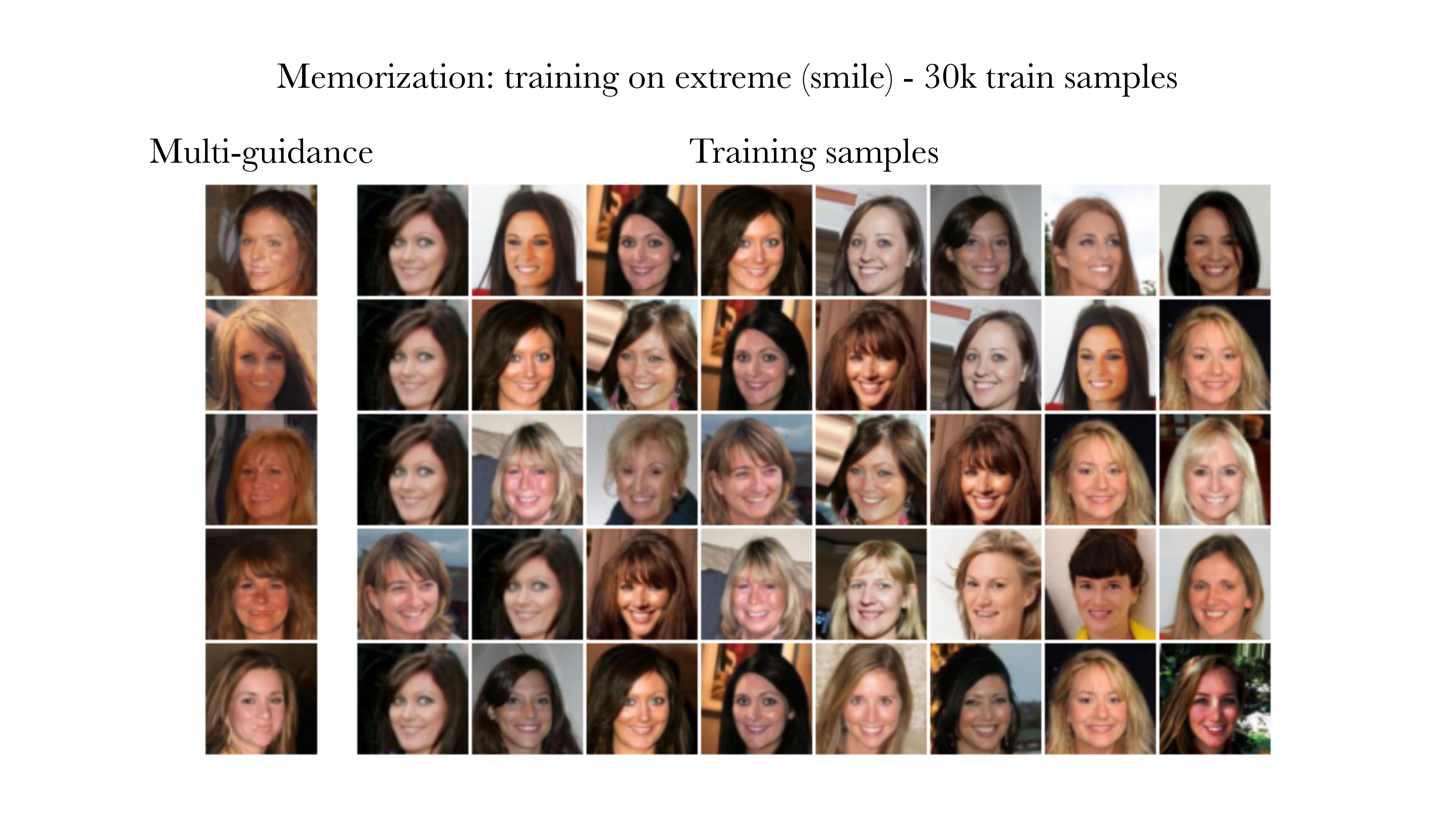}
    \caption{Checking for memorization when training on $30$k extreme examples from CelebA. The generated samples are on the left, while the right side showcases the nearest neighbors according to CLIP embeddings.}
    \label{fig:smile-30k-memorization}
\end{figure}

\begin{figure}[t]
    \centering
    \includegraphics[width=1\linewidth]{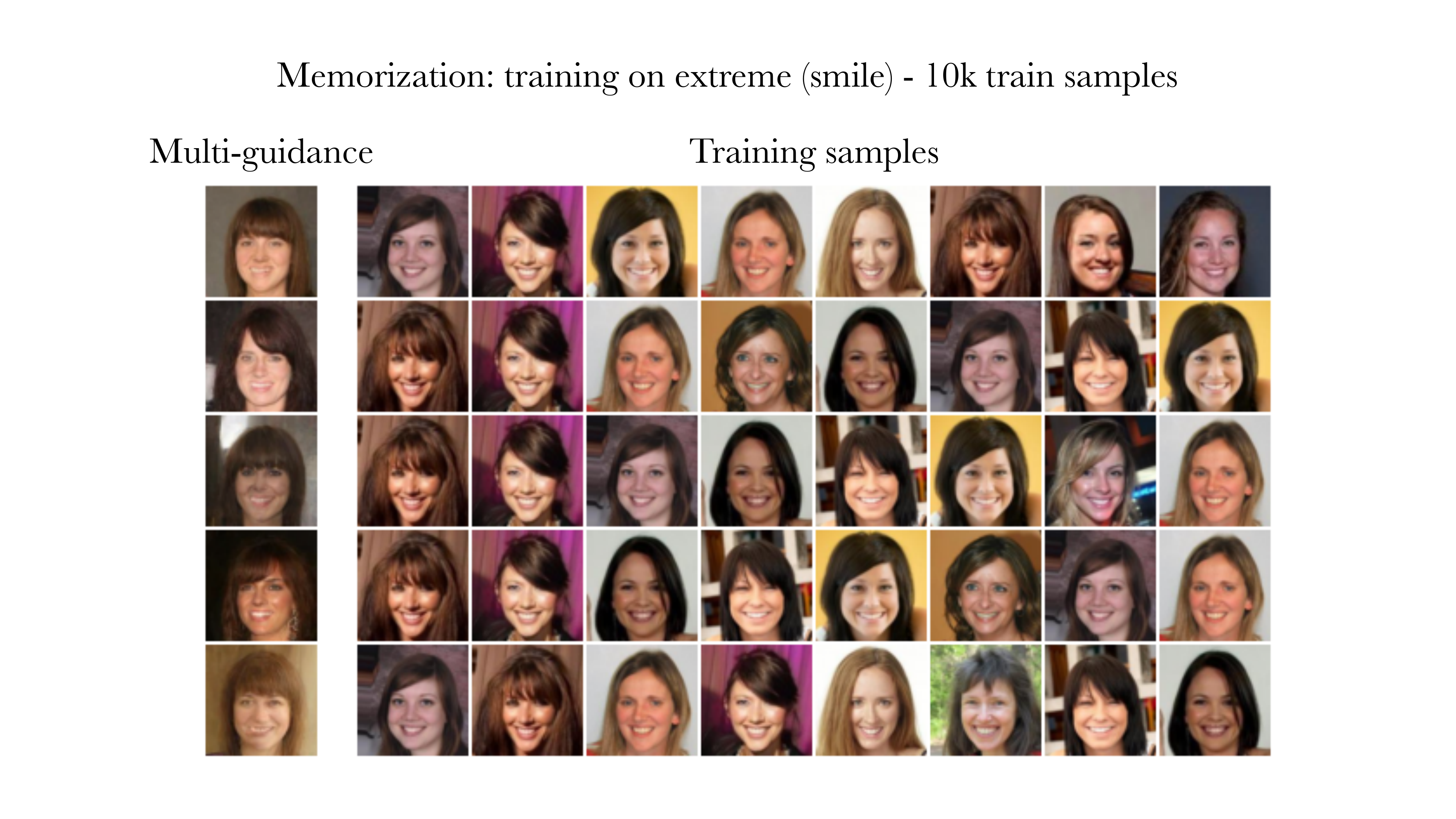}
    \caption{Checking for memorization when training on $10$k extreme examples from CelebA. The generated samples are on the left, while the right side showcases the nearest neighbors according to CLIP embeddings.}
    \label{fig:smile-10k-memorization}
\end{figure}

\begin{figure}[t]
    \centering
    \includegraphics[width=1\linewidth]{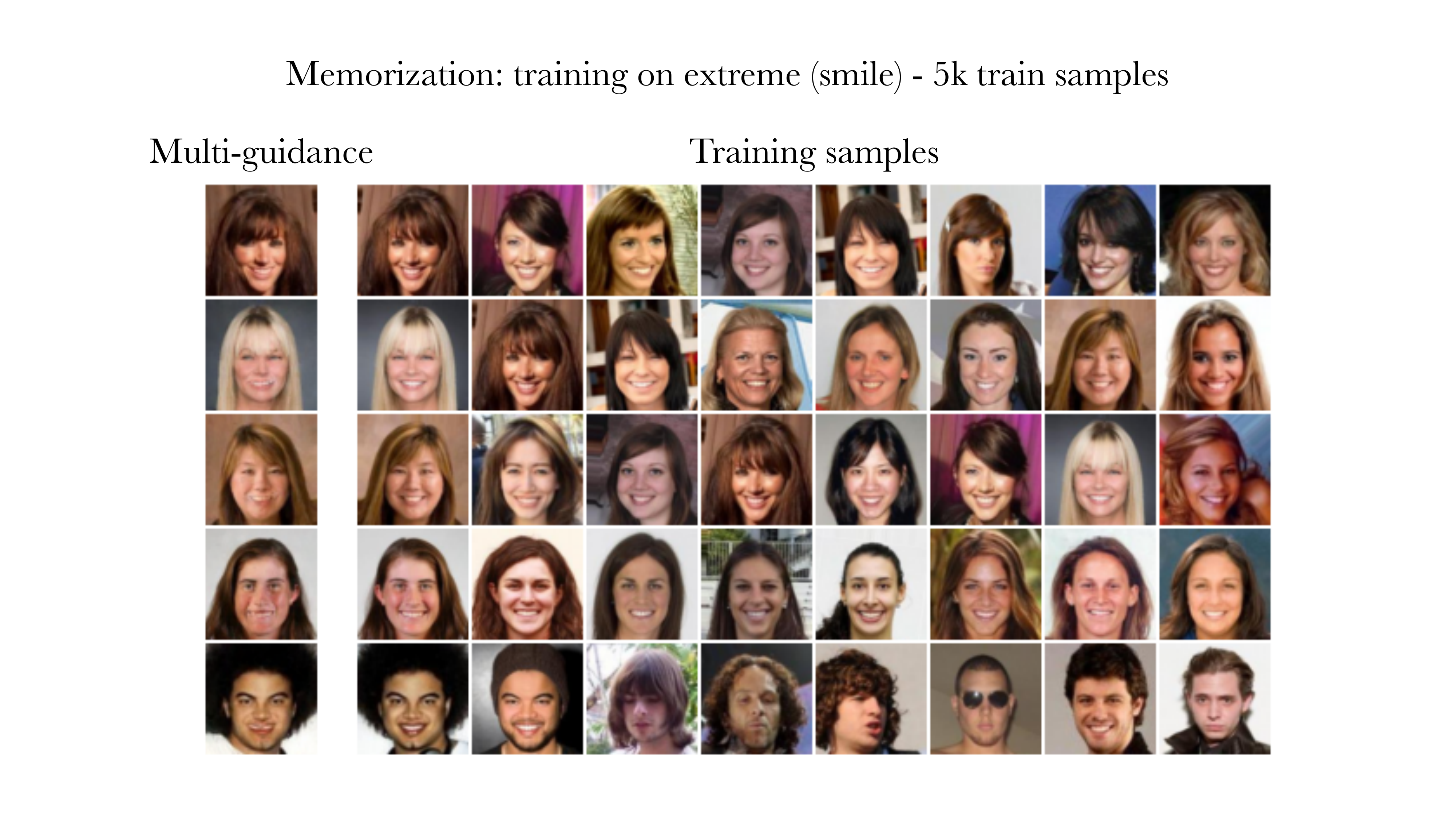}
    \caption{Checking for memorization when training on $5$k extreme examples from CelebA. The generated samples are on the left, while the right side showcases the nearest neighbors according to CLIP embeddings.}
    \label{fig:smile-5k-memorization}
\end{figure}

As memorization has been observed in diffusion models \citep{somepalli2022diffusion}, we rigorously examine whether our DDPMs memorize the training data, especially when reducing the size of the training set. In the figures below, we present synthetic examples with their nearest neighbor based on CLIP embeddings \citep{radford2021learning}. While models trained on $5$k images exhibit memorization, increasing the number of images mitigates this effect. Refer to Figures \ref{fig:smile-60k-memorization}, \ref{fig:smile-30k-memorization}, \ref{fig:smile-10k-memorization}, \ref{fig:smile-5k-memorization} for the examples with nearest neighbors in the training data for different data sizes.
\begin{table}
    \centering
    \caption{\rebuttal{FID, Precision (P) and Recall (R) for different guidance parameters. The columns with "full" contain the metric computed between multi-guidance samples and $50k$ random samples from the training data. The columns with "mild" contain the metric computed between multi-guidance samples and the $10k$ training samples with predicted smile attribute closest to $0.5$ in $\ell_1$ distance.}}
    \label{tab:fid-precision-recall}
    \begin{tabular}{|c|c|c|c|c|c|c|}
        \hline
        Guid. strength & FID (mild) & FID (full) & P (mild)  & P (full)  & R (mild)  & R (full)  \\
        \hline
        0.0    & \textbf{12.7215} & \textbf{8.1421}  &         0.7564  &         0.7876  & \textbf{0.4755} &          0.484   \\
        3.5    &         13.8471 &          10.17    &         0.8064  &         0.8164  &         0.4143  &          0.4190  \\
        10     &         14.4841 &          10.72    & \textbf{0.8182} & \textbf{0.8195} &         0.4135  &          0.4148  \\
        30     &         14.9275 &          11.34    &         0.8018  &         0.8051  &         0.4353  &          0.4250  \\
        50     &         14.6142 &          12.06    &         0.7756  &         0.7797  &         0.4365  &          0.4275  \\
        75     &         14.7289 &          13.11    &         0.7349  &         0.7317  &         0.4328  &          0.4225  \\
        100    &         15.1242 &          13.58    &         0.6549  &         0.6623  &         0.4713  &  \textbf{0.5383} \\
        \hline
    \end{tabular}
\end{table}
\section{Measuring Sample Quality and Diversity}
\label{sec:measuring-quality-and-diversity}
\rebuttal{We evaluate synthetic image quality and diversity using the FID \citep{ttur-gan-fid} and Precision and Recall \citep{sajjadi2018assessing} metrics. We compare synthetic images against two sets of real images: $50$k randomly selected samples from the CelebA training data and $10$k examples with continuous labels nearest to 0.5 under the $\ell_1$ distance. These subsets allow us to assess how well multi-guidance sampling captures both the entire data distribution and the section containing mild examples only. Results for different values of $\lambda$ are presented in Table \ref{tab:fid-precision-recall}.}
\paragraph{FID} 
\rebuttal{The trends in FID scores align with our manual inspection, showing increasingly frequent artifacts in samples generated with multi-guidance for larger values of $\lambda$. Note that our FID values are higher than those typically reported in state-of-the-art diffusion models. This discrepancy can be attributed to our shorter training runs for computational efficiency. Our primary goal is to demonstrate interpolation behavior rather than to surpass the state-of-the-art in diffusion models.}
\paragraph{Precision and Recall}
\rebuttal{We observe that Precision and Recall correlates less with our manual assessment. Indeed, multi-guidance sampling captures mild smiles better than unconditional sampling, leading us to expect higher recall for multi-guidance, particularly when comparing synthetic samples with the most mild ones. Regarding Precision, we were surprised to find that computing precision for either the entire distribution or just the mild part resulted in similar trends. These findings suggest that the embedding model used to compute the Precision and Recall, the Inception network of \citet{szegedy2015rethinking}, might struggle to capture the nuances of our problem. Throughout the paper, we primarily use $\lambda = 30$ for guidance, as we found it to introduces minimal artifacts while effectively generating mild attribute examples.}
\section{Interpolation with Multi-Guidance under Additional Settings}
\label{sec:interpolation-with-different-hyperparameters}
\begin{figure}[t] 
    \centering
    \subfloat[]{
        \includegraphics[width=0.45\linewidth]{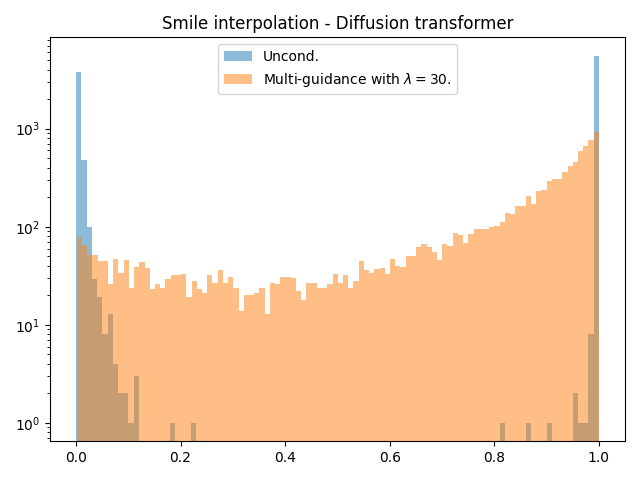}
        \label{fig:multi-guidance-dit}
    }
    \hfill
    \subfloat[]{
        \includegraphics[width=0.45\linewidth]{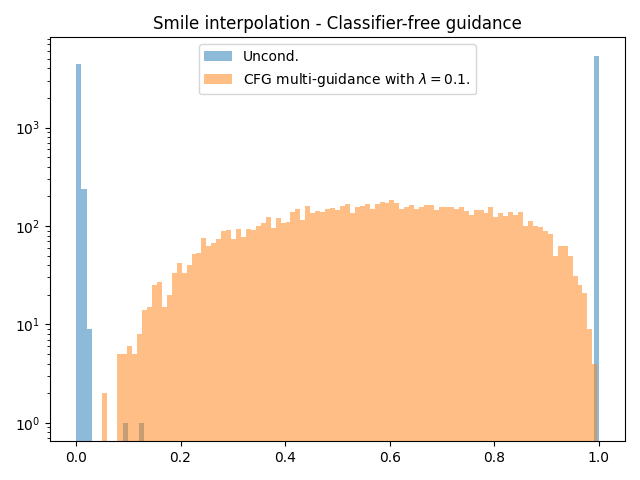}
        \label{fig:multi-guidance-cfg}
    }
    \caption{Distribution histogram of smile sampled with (a): a diffusion transformer instead of a U-net; (b): when sampling with classifier-free guidance.}
    \label{fig:interpolation-dit-cfg}
\end{figure}
\rebuttal{We observe interpolation behavior in diffusion models by varying the architecture, as well as when sampling with classifier-free guidance \citep{ho2022classifierfree}. Additionally, we attempted to train a DDPM using the $\mathcal L_\text{VLB}$ objective, but the model did not fit the training distribution well; in other words, unconditional sampling did not produce realistic-looking samples. Due to our limited computational budget, we did not extensively tune hyperparameters to successfully train a model using $\mathcal L_\text{VLB}$. Refer to \cref{fig:interpolation-dit-cfg} for histograms.}
\rebuttal{\paragraph{Alternative architecture} Instead of using the U-net architecture proposed by \citet{improved-ddpm}, we replaced it with the diffusion transformer introduced by \citet{peebles2023scalable}. As shown in \cref{fig:multi-guidance-dit}, diffusion models implemented with a transformer also demonstrate interpolation capabilities. This suggests that the ability to interpolate does not depend on the choice of architecture.}
\paragraph{Sampling with Classifier-Free Guidance}
\rebuttal{
Instead of classifier guidance \citep{dhariwal2021diffusion}, one can sample from a diffusion model using classifier-free guidance \citep{ho2022classifierfree}. When adapting multi-guidance to classifier-free guidance, we still observe interpolation capabilities of the diffusion model. Interestingly, the shape of the distribution when sampling with classifier-free guidance differs from samples generated with classifier guidance. Specifically, there is little-to-no mass on the edges of the histogram. See \cref{fig:multi-guidance-cfg} for the histogram. We adapt the classifier-free sampling algorithm by using the following score for sampling:
}
\begin{equation}
    s_\text{cfg} (x) = - \lambda s_\theta(x, \varnothing) + (1 - \lambda) \frac{1}{2} (s_\theta(x, 0) + s_\theta(x, 1))
\end{equation}
\rebuttal{where $s_\theta(x, i)$ is the score learned by the model for the class $i$, and $s_\theta(x, \varnothing)$ denotes the unconditional score.}
\rebuttal{\paragraph{Noise schedule and diffusion variance} We observed that interpolation occurs with both the cosine and linear noise schedules from \citep{improved-ddpm}, as well as with fixed and learned variance of the diffusion process.}
\section{Demonstrating Interpolation on Objects of Different Sizes}
\label{sec:object-size-experiment}
\begin{figure}[t] 
    \centering
    \subfloat[]{
        \includegraphics[width=0.18\linewidth]{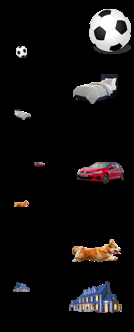}
    }
    \hfill
    \subfloat[]{
        \includegraphics[width=0.78\linewidth]{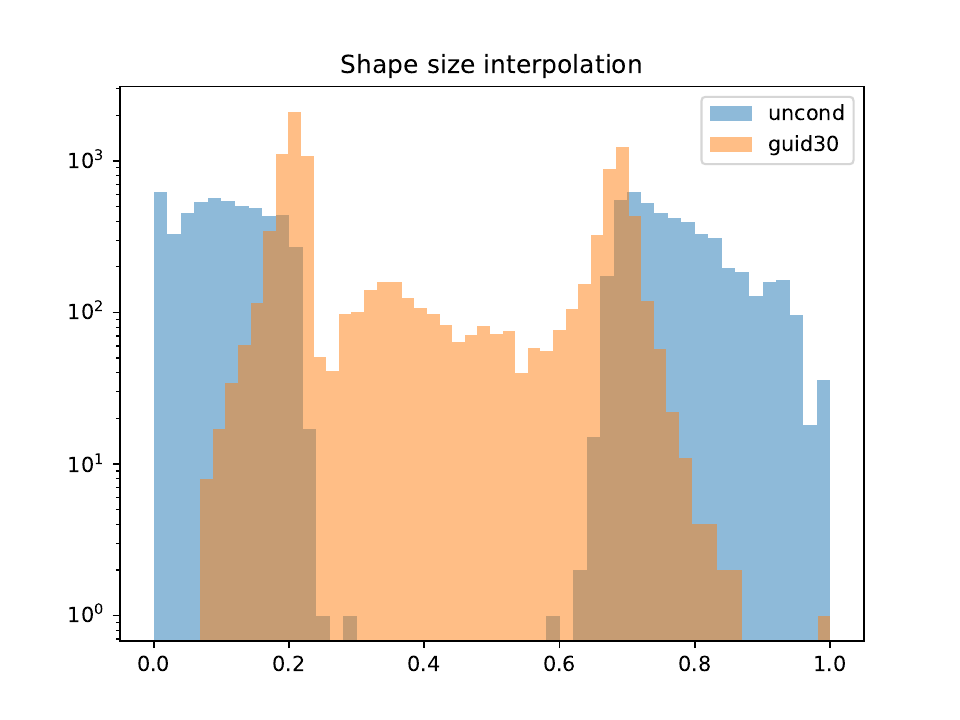}
    }
    \caption{Left: training examples with rescaled real-world objects. Right: distribution of object sizes when sampling from the unconditional model or with multi-guidance.}
    \label{fig:object-shape-experiment}
\end{figure}
\rebuttal{In addition to the experiments detailed in the main body, we conducted an experiment to measure interpolation using a dataset of real-world objects resized to various dimensions. In this experiment, we define extreme examples as pictures with objects of the 20\% smallest and largest sizes. Examples from the training data as well as histograms are displayed in \cref{fig:object-shape-experiment}. Since we know the ground-truth object size in this experiment, we train the evaluation classifier to regress on the continuous attribute in the range $[0, 1]$ instead of training it on discretized labels.}
\section{Multi-Guidance on ImageNet}
\label{sec:multi-guidance-imagenet}
\begin{figure}[t] 
    \centering
    \includegraphics[width=0.25\linewidth]{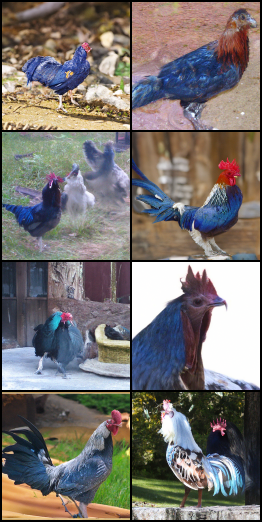}
    \caption{Samples generated with multi-guidance between the classes "Hen" and "Jay". While we believe that interpolation between thoses classes is not well-defined, we include this result as it may interest the reader.}
    \label{fig:blue-chicken}
\end{figure}
\rebuttal{Sampling with multi-guidance from a model trained on ImageNet using the scores for the classes "Jay" and "Hen" produces images of blue chickens. Examples can be viewed in \cref{fig:blue-chicken}. This result is interesting, but we caution that it is not immediately clear that there exists a continuous range from "Hen" to "Jay". Nonetheless, we decided to share this since reviewers were curious about experiments on larger datasets such as ImageNet.}
\section{Interpolation between DDIM Latents}
\label{sec:interpolation-ddim-latents}
\begin{figure}[t] 
    \centering
    \includegraphics[width=1.0\linewidth]{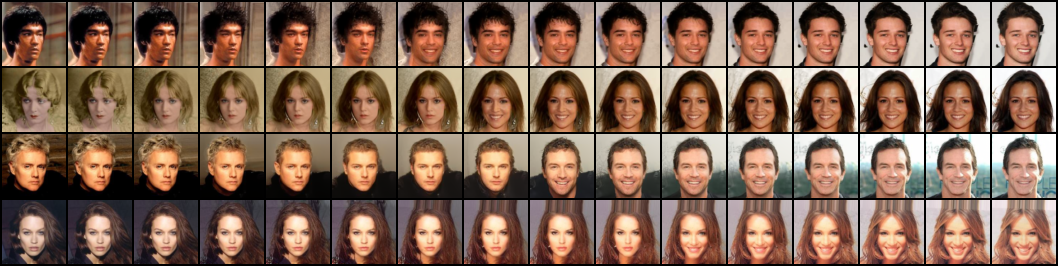}
    \caption{Spherical linear interpolation (slerp) between ddim latents. We see that intermediate regions of the latent space get decoded as mild smiles by the diffusion model trained solely on extreme examples.}
    \label{fig:ddim-interpolation}
\end{figure}
\rebuttal{Building upon prior work on generative models such as \citet{radford2016unsupervised, upchurch2017deep, berthelot2018understanding, kingma2018glow, karras2019style, harkonen2020ganspace}, we study interpolation in the noisy latent space of diffusion models trained on extreme examples only. As for the rest of this paper, this investigation is crucial as prior research either trains models on the whole data distribution or relies on embeddings from multi-modal models trained on internet-scale datasets \citep{song2022denoising, wang2023interpolating}.}
\rebuttal{Similar to DDIM \citep{song2022denoising}, we define the latent space of diffusion models as the partially denoised samples, i.e. samples at non-zero noise levels. Interpolation within the latent space is achieved through a multi-step process: we first compute the latent representations of two real images after injecting noise for 2000 steps, then perform spherical linear interpolation (slerp), and finally, apply deterministic denoising using DDIM without noise injection. Although we experimented with regular linear interpolation, we found that slerp consistently achieves superior results. \cref{fig:ddim-interpolation} displays interpolation results between pairs of samples.}
\section{Attribute Editing without Training on Mild Examples}
\label{sec:attr-edit-without-mild}
\begin{figure}[t] 
    \centering
    \includegraphics[width=1.0\linewidth]{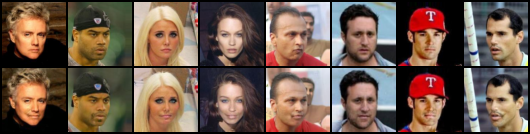}
    \caption{Attribute edition using classifier guidance. Given a real image, we inject noise and decoding with regular classifier guidance yields images with mild attributes.}
    \label{fig:smile-editing}
\end{figure}
\rebuttal{Following discussions with reviewers during the rebuttal phase, we conducted experiments on attribute editing using diffusion models trained exclusively on extreme examples. While previous work has shown successful interpolation in unconstrained settings \citep{kim2022diffusionclip, kong2022leveraging, gandikota2023unified, yue2024exploring}, we demonstrate that even when trained solely on extreme examples, diffusion models can still exhibit interpolation capabilities. Specifically, we use the same DDPM model from in the interpolation experiments detailed in \cref{sec:smile-interpolation-subsec}.}
\rebuttal{We perform attribute editing by first mapping the original image to DDPM latents using 2000 reversing steps (half the number used during training). Subsequently, we denoise using regular classifier guidance with a guidance strength of 30. We use DDIM with noise injection using a linearly decaying schedule $(\eta_t)_{t=1}^{2000}$ such that $\eta_t = \frac{t}{2000}$. For further details on $\eta_t$, please refer to the original DDIM paper \citep{song2022denoising}.}
\rebuttal{Using this straightforward approach, we observed that DDPMs could transform non-smiling faces into mildly smiling ones. CelebA images with their "smile" attribute edited can be seen in \cref{fig:smile-editing}.}